\def\year{2019}\relax
\documentclass[letterpaper]{article}
\usepackage{aaai19}
\usepackage{times}
\usepackage{helvet}
\usepackage{courier}
\usepackage{url}
\usepackage{graphicx}
\usepackage{algorithm}
\usepackage{algorithmic}
\usepackage{amsfonts}
\usepackage{amssymb}
\usepackage{amsmath}
\usepackage{bm}
\usepackage{booktabs}
\usepackage{subfig}
\usepackage[title]{appendix}
\title{Rethinking of AlphaStar}
\author{Ruo-Ze Liu\\
	liuruoze@163.com
}

\begin{document}

\maketitle

\begin{abstract}
	We present a different view for AlphaStar (AS), the program achieving Grand-Master level in the game StarCraft II. It is considered big progress for AI research. However, in this paper, we present problems with the AS, some of which are the defects of it, and some of which are important details that are neglected in its article. These problems arise two questions. One is that what can we get from the built of AS? The other is that does the battle between it with humans fair? After the discussion, we present the future research directions for these problems. Our study is based on a reproduction code of the AS, and the codes are available online .
\end{abstract}

\section{Introduction}
For a long time, the board game Go was considered a testbed challenging for AI technologies. Its state space is huge (about $3^{361}$) thus traditional searching methods can not solve it in a reasonable time. AlphaGo (AG)~\cite{silver2016AlphaGo}, a program released by DeepMind (DM), beat the human Go Champion player in 2015, by using deep reinforcement learning (Deep-RL, DRL)~\cite{Mnih2015DQN} and Monte Carlo tree search (MCTS)~\cite{Chaslot2008MCTS}. AG first used human data to do supervised learning (SL) and then used DRL plus MCTS to do RL training~\cite{sutton1998RL}. After 1 year, the successor, AlphaGo Zero (AG$^Z$)~\cite{silver2017AlphaGoZero} was published, which can beat AG without human data. Researchers thought Go has been solved and began to shift to other challenging domains.

Meanwhile, from 2010, a video game StarCraft (SC, or called SC1) developed by Blizzard Entertainment (BE) was also used as a testbed for AI~\cite{ontanon2013SCsurvey}. SC has many differences with Go. E.g., it is an imperfect information game, its action and state space are much larger. Though there had been some RL research on SC, the majority of methods are state machines and heuristic search, e.g., see SAIDA~\cite{SAIDA} and Locutus~\cite{Locutus}. These bots' APM (actions per minute) is not restricted, meaning their APM can be much larger than humans'. However, there are still no bots that can beat professional SC players~\cite{Certick2019starCraftAIcompetition}.

The two roads of RL research after Go and AI research for SC converged in 2017. DM and BE jointly published SC2LE (short as SE~\cite{vinyals2017sc2}), a machine learning platform for StarCraft II (SC2). SC2 is the successor to SC, also developed by BE. The gameplay of SC2 is similar to SC1 except for some differences like unit types have changed a little. Such, training difficulties on the two should be similar. Previously, a learning platform TorchCraft~\cite{Synnaeve2016torchcraft} existed for SC1. However, SE is a better one due to 3 reasons: 1. BE provided an official interface for SC2, while the one in SC1, the BWAPI, is provided by the community through reverse engineering; 2. BE also released Linux SC2 to be the simulator for running it on servers. While SC1 can not directly run on Linux, making training inconvenient; 3. BWAPI can't close the game rendering screen, leading to training slow. While the Linux SC2 can run without rendering, improving the sampling speed by a large margin.

% got through reverse engineering

DM also provided supports for SE by a Python interface PySC2$_1$ (short as PS$_1$, the subscript is version number). A contribution of PS is that it provides a human interface (HI) for agent learning. E.g., the agent needs first to select a unit (like humans use a mouse to click), then issues the command, which we call human action (HA). The agent also needs to perceive visual information directly from the screen. Such challenges make training on PS is hard. The baseline~\cite{Mnih2016A3C} can not even beat the easiest built-in AI. Some researchers used tricks to ease the difficulties, e.g., see~\cite{sun2018tsarbot},~\cite{pang2019sc2}, and~\cite{Lee2018sc2}. They rely on macro actions built on HA. However, direct training based on HA is still little. 

%Note, for SC1, there is only a raw interface (RI).
	
In Jan 2019, AlphaStar (AS)~\cite{Vinyals2019ASBlog} was proposed in a blog (we call it AS$^B$, B refers to Blog). When fighting vs. human players (HP), AS$^B$ showed powerful performance and beat two professional HPs. However, in a Live show, AS was deceived by the strategies of an HP, showing its weakness. Also, some criticisms argue it uses high APM humans can not achieve. After 10 months, the improved AS~\cite{Vinyals2019ASNature} was published (we call it AS$^N$, N refers to Nature). AS$^N$ has a restriction of APM and uses a camera. However, AS$^N$ still achieved the Grand-Master (GM) level in BattleNet (BN, a match platform hold by the game developer) and surpassed $99.8\%$ of players. The results are amazing and the SC2 problem seems to be solved.
	
However, in this paper, we present a different view of AS by delving into its details, codes (by a mini-scale reproduction~\cite{liu2021mAS}), experiments, and analysis of replays (90 ones). We find that the problem handled by AS is actually a sub-problem of the one proposed in SE. The difficulty between the two is much different but is little discussed. We call this ``Neglect in AS". Then through replays, we present the defects of AS and connect the reason to its architecture. The two follow-ups of AS are also discussed. Finally, we give future directions for the SC2 problem. This paper's contributions are as follows:
\begin{itemize}
    \item We present a critical rethinking of AS, such as the actual problem it handled and the defects of it.
    \item A symbolic definition for the SC2 problem and its decomposition on several dimensions are given.
    \item For future research on SC2 AI, we give useful directions.
\end{itemize}

\section{Background}

\subsection{Deep Reinforcement Learning}
RL can be reprsented as a Markov decision procsee, which is a 6-tuple $\left\langle S, A, P, R, \gamma, T \right\rangle $. For an environment, $ S $ is the state space and $ A $ is the action space. From time step $0$ to $T$, the agent faces with state $s_t \in S$, and selects action $a_t \in A$ by policy $\pi$ (i.e., $a_t \sim \pi(s_t)$). Then the agent receives a reward $r_t = R(s_t, a_t)$ and the environment transists to next state $s_{t+1} \sim P(s' | s_t, a_t) $. $ \gamma $ is a discount factor and return is $G_t = \sum_{z=t}^T\gamma^{z-t} r_z $. An RL algorithm aims to find the optimal policy $\pi^* = \text{argmax}_\pi \mathbb{E}_t[G_t] $. 

If using function approximation (FA) to represent the policy and the FA is a Deep Neural Network (Deep-NN, or DNN)~\cite{LeCun2015DL}, such RL is called DRL. There have several types of DNN in DRL. E.g., CNN (Convolutional NN)~\cite{Lecun1998CNN,Krizhevsky2012AlexNet} is for image-like data learning. RNN (Recurrent NN or LSTM)~\cite{Hochreiter1997LSTM} is good of processing sequence data. Transformer~\cite{Vaswani2017Transformer} is a fully connected NN with a self-attention mechanism which is good at handling structural information. The learning abilities of DRL benefit from the representing power of these DNNs.

%The learning abilities of DRL benefit from the representing power of these DNNs.

\subsection{Problem of StarCraft II}
\begin{figure}[ht]
    \centering
    \includegraphics[width=0.85\columnwidth]{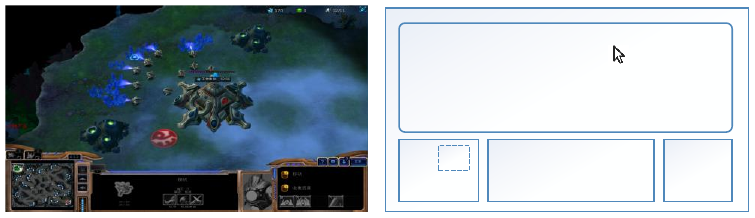}
    \caption{Left: screenshot. Right: schematic. The top is a screen, the bottom left is a minimap, the bottom middle is a user interface, dashed box in the minimap is the camera. The mouse can be clicked on the screen and minimap.}
    \label{fig:Screen}
\end{figure}

In SC2, the player chooses one from $4$ race settings (Protoss, Terran, Zerg, and Random) to play. At the start, the $2$ players are born in diagonal locations on the game map. Players need to collect resources, build buildings, produce units, and destroy all the opponent's buildings to win. An illustration is in Fig.~\ref{fig:Screen}. Several features make SC2 suitable for testing AI, such as imperfect information, various types of units, and requirements for operation speed (more details in Appx.~\ref{appen:more of SC2}). We formulate the problem of SC2 as $\mathbb{SC}$ and later we will show how to divide it into several dimensions. 

% The effective methods on SC2 show more practical value, because SC2 is similar in complexity as the real-life task. 

%The effective methods on SC2 show more practical value, because SC2 is more similar in complexity as the real-life task. We formulate the problem of SC2 as $\mathbb{SC}$.

%SC2 is a heavy game (heavy means, compared to Go, it needs more resource for running, training, and debugging). Such, SC2 is not an ideal research environment. However, due to this, the effective methods on SC2 show more practical value, because SC2 is more similar in complexity as the real-life task. We formulate the problem of SC2 as $\mathbb{SC}$. 

%  (note destroying all units is not necessary)

\subsection{AlphaStar Blog}
The baseline in SE (we call it AS$^Z$, Z refers to zero) uses a similar DNN (CNN and LSTM) and RL algorithm (A3C) as in AS$^B$. But why it has a large improvement in AS$^B$? One reason is $\mathbb{SC}$ faced by AS$^B$ has changed compared to AS$^Z$. 

\begin{figure}[ht]
    \centering
    \includegraphics[width=1.00\columnwidth]{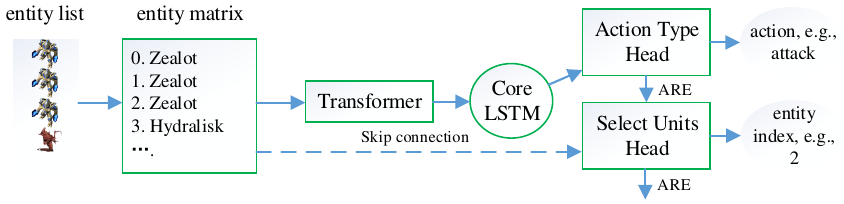}
    \caption{How from an entity list to an index on AS$^{B\&N}$.}
    \label{fig:entity}
\end{figure}

AS$^Z$ used HI while AS$^B$ uses the raw interface (RI). The RI provides an entity list through the game engine (GE) which includes units and buildings of ours, enemies, and neutral's. By that, a selection of an entity is replaced by passing an identifier to it. The identifier in the GE is \textit{tag} (a unique number in one episode of game, e.g., like ``8942387201''). AS$^B$ first turns the entity list to an entity matrix and then uses a Transformer to model entities' relationship (note the row number in the matrix equals the index in the list). Then, in the selected units head, the entity that performs the action is identified by AS$^B$ using the index (see Fig.~\ref{fig:entity}). AS$^B$ calls RI to transform the index to the entity's tag value and calls the GE to use the tag to perform the actions. In this way, AS has no need to select a unit like humans, or to remember the tag value. Thus the training difficulty is reduced by a margin.

%  (the same as unit, ``unit'' refers to unit or building according to context) 

% The list is given as the ``raw units'' interface in the PS (we later discuss the version). 
%Some actions need a target unit. Note the entity list also includes enemies and neutral's. Hence the target's tag is also translated by the index. 
%We can see, through the raw units and raw actions, and the added Transformer, AS$^B$ successfully builds the relationship from the entity to actions. Its action doesn't need the separation of selection and commands.

% (based on PS$_3$, later we will discuss the versions of PS)

\subsection{AlphaStar Nature}
AS$^N$ architecture consists of 3 encoders, 1 core, and 6 heads. The 3 encoders correspond to 3 types of information: spatial, statistical, and entity. The spatial one corresponds to minimap sub-layers passed into ResNet~\cite{He2016ResNet} (one type of CNN, note screen information is not used here). The statistical one includes minerals, food, and so on. The entity information is not used in AS$^Z$ but in AS$^{B\&N}$ (means AS$^B$ and AS$^N$). The outputs of encoders pass into the core LSTM module of which the outputs pass into an action-type head. Besides the action type, this head also outputs an auto-regressive embedding (ARE). ARE passes into the next argument module. This is repeated for all 5 modules: delay, queued, selected units, target unit, and target location.

For sample efficiency, AS$^N$ uses an off-policy RL algorithm. Its loss includes V-trace actor-critic~\cite{Espeholt2018VTrace} loss, UPGO loss, entropy loss, and distillation loss between the SL policy and the RL one. The tactics in SC2 form a cycle. AS uses a league training mechanism to train a robust policy. But, we will show that AS is still easy to be attacked by novel tactics. Note AS$^N$ trains 1 agent for 1 Race (AS$^N_P$ refers to Protoss). AS$^N$ has 3 trained versions. We only discuss the Final version here.

%AS$^N$ has 3 trained versions: Supervised, Mid, and Final. We only discuss the Final version, here.

%Also, its detection ability is not satisfied.
\section{Neglect in AlphaStar}
We present the 5 important details that are neglected in AS.
%Here we present why the $\mathbb{SC}$ handled by AS$^{B\&N}$ is not the $\mathbb{SC}$ proposed in SE.

%We now analyze the details which are important but are neglected in AS's paper. These details reveal why the problem handled by AS is not the original one proposed in SE. 

\subsection{Human Interface vs. Raw Interface}
When SE was first proposed, there was a vision that $\mathbb{SC}$ faced by RL should be much different from other domains. The $\mathbb{SC}$ should be handled as similar as humans, that is, using HI to learn. HI consists of HA and HO (human observation), as $ \text{HI} = \text{HA} + \text{HO}$. E.g., the agent should select a unit by its coordinate and only use the image on the screen to infer the information. At that time, the community agreed to that consensus and researches have been made(e.g.,~\cite{pang2019sc2}). We call this ``human-level control on SC2".

However, to the surprise of some people, AS$^{B\&N}$ used the RI instead of the HI to build its agent, which cast a cloud over its achievements. Note, $ \text{RI} = \text{RA} + \text{RO} $, where RA is the raw action the GE uses, RO is the raw observation the GE sees (e.g., the entity's information). RI makes AS's achievements not as great as the vision proposed in SE. It not only narrows the generality to other domains but also makes the comparison with humans not so fair (we later will show it). If we formulate $\mathbb{SC}$ in SE as $\mathbb{SC}^h$ (h refers to human), the $\mathbb{SC}$ handled by AS is actually $\mathbb{SC}^r$ (r refers to raw). 

\begin{figure}[ht]
    \centering
    \includegraphics[width=1.00\columnwidth]{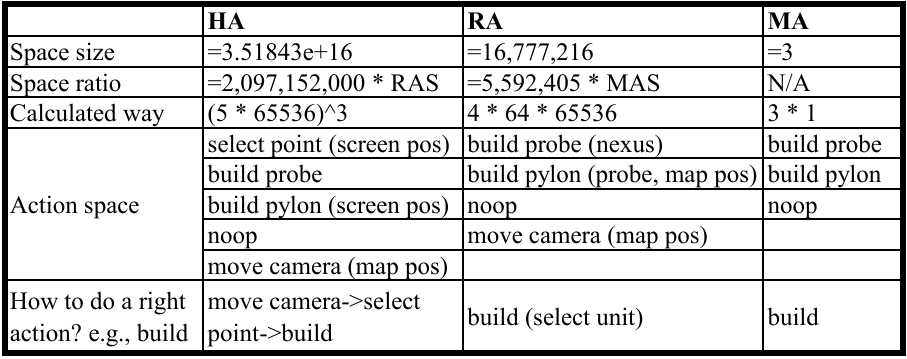}
    \caption{The $3$ different action spaces for a prototype problem. HA=human action. RA=raw action. MA=macro action. Size of entity list is 64. Map size is 256x256.}
    \label{fig:rawvshuman}
\end{figure}

We present a case showing the learning difference between RA and HA. Fig.~\ref{fig:rawvshuman} shows the action space size of the two ones. This is a simple prototype in SC2, that is, how to use Protoss to get as much economy as possible at the start of the game. Through experiments, we find the training on HA is much more difficult than on RA (the training of RI is faster, while the curve of HI is flat. We place the details and training curve in Appx.~\ref{appen:experi of ravsha}), which verified the analysis.

% The reproducible source codes are presented in the code appendix

% \begin{figure}[h]
%     \centering
%     \includegraphics[width=0.90\columnwidth]{./RvH.pdf}
%     \caption{Training curves on the two action spaces.}
%     \label{fig:rawvshuman2}
% \end{figure}

Note, the difference between RI and HI is important but is little discussed in AS$^B$~\cite{Vinyals2019ASBlog} which only states that it uses the ``raw interface'' from PS. However, the PS$_2$ at that time doesn't include RA, which makes people suspect that the ``raw" refers to the original interface (the HA). Only when the AS$^N$ is published and PS$_3$ is available, the real meaning of ``raw'' is revealed (PS$_3$ is released at the time near as AS$^N$, see Table~\ref{tab:PS versions}). The PS$_3$ contained the RA now. But, its RA is not a replacement or upgrade on the original HA, but a neighbor. So PS$_3$ has 2 action spaces now. It also adds similar functions for using RA as the functions using HA, which makes things look like the RA is actually added as needed afterward.

\begin{table}[ht]
    \centering
    \scalebox{1.0}{
    \begin{tabular}{l | c c c c  }
    \toprule
    Type  & RO & RA & Date & Works on \\
    \midrule
    PS$_1$  & No & No & Aug 2017 & AS$^Z$ \\
    PS$_2$  &  Yes  & No & Jun 2018 & -    \\
    PS$_3$  &  Yes  & Yes & Sep 2019 & AS$^{B\&N}$   \\
    \bottomrule
    \end{tabular}
    }
    \caption{PS versions. PS=PySC2. AS=AlphaStar. RO=raw observation. RA=raw action.}
    \label{tab:PS versions}
\end{table}

AS$^N$~\cite{Vinyals2019ASNature} gives a statement of the RI but neglects the discussions of the differences between RI and HI. We argue that these discussions should be careful and comprehensive, otherwise it could make people think $\mathbb{SC}$ in SE is already solved. That is, people would see no future research on SC2 AI is needed, which could be very harmful to the community of the SC AI research. 

\subsection{Too Many Human Knowledge}
Previously, AG$^Z$ uses no human data for training while AG uses human replays to learn an initial policy. Differently, AS uses human data in $3$ aspects. The 1st is the same as in AG, which uses human replays to learn an SL policy. 

The 2nd is the z reward used in RL which are statistic values that consist of two parts, the build order and unit counts (we place their meaning in Appx.~\ref{appen:term of SC2}). The statistics may be important for SC2, but may not be so important for other domains, which narrows the AS's generality. The usage of z is as below. Before each running of the game, AS selects a replay from the human replays. The race and map of the replay should be equal to the one in the current game. Then, when the agent runs in the SC2 environment, the replay also runs for the same steps. In each step, AS calculates the difference of the z between the agent and the player in the replay, transforming it as a negative reward. The reward is then used for RL learning. Note the z reward makes AS learning improved, but it also promotes AS to imitate the human player, instead of encouraging it to explore new tactics.

%  
% That is, the less difference the larger the reward. 

The 3rd aspect is the distillation loss of which the target is the SL policy. This loss makes the learned RL policy not moving far away from the SL one. It has the pros to stable training yet cons of preventing the agent from exploring. 

%  new strategies

There have some other tricks in AS$^N$ using human data. A typical one is to use the unit type to decide the range of the action type. Note PS doesn't provide such information. Thus this must be coded manually which is another way of injecting human knowledge in AS. 

% HINT: can be reduced
If we formulate $\mathbb{SC}$ using no human knowledge as $\mathbb{SC}_0$. Then the one using human replays is $\mathbb{SC}_1$. The way AS uses is actually $\mathbb{SC}_3$ (meaning it uses human data in $3$ ways). The ideal problem we want to solve is $\mathbb{SC}_0$. The target problem we would solve is $\mathbb{SC}_1$. Actually, there is a gap from $\mathbb{SC}_3$ to $\mathbb{SC}_1$ and a huge gap from $\mathbb{SC}_1$ to $\mathbb{SC}_0$ (inferred by the difference between AG and AG$^Z$). 
 
\subsection{Impact of APM and EPM}
In the match against professional players, AS$^B$ showed incredible APMs to be criticized. In the AS$^N$, DM restricted its APM. But there are still some things to be noted.

We now introduce EPM, a metric for measuring the player has how many ``effective actions per minute". The ``effective" means useful, e.g., a click on one unit has meaning, but click it twice or third times has no meaning because latter clicks change no state. For humans, they will perform many unuseful clicks, which is to warm up their hands to prepare for the later fierce battle. This will generate many unuseful actions. In order to distinguish them, SC2 already provides the EPM to measure them. The EPM only contains useful actions, removing unuseful ones. People can switch seeing the APM or EPM of the player in a replay.

% E.g., the key combat may happen from 5:00 to 10:00, but the players will warm their hands through 0:00 to 5:00. 
% The unit is already selected, clicking the unit will only refresh the select status (this status is forever), so more clicking is meaningless.

\begin{table}[ht]
    \centering
    \scalebox{1.0}{
    \begin{tabular}{l | c c | c c }
    \toprule
    Type  & EPM$^A$ & EPM$^P$  & APM$^A$ & APM$^P$ \\
    \midrule
    Value  &   \textbf{182.10}    &  154.17    &  200.17 &  \textbf{247.28}   \\
    \bottomrule
    \end{tabular}
    }
    \caption{Average APM and EPM for all AS$^N_P$ replays. Superscript A=AS, P=Player. Bolds are larger values. }
    \label{tab:epm and apm}
\end{table}

We give an example that showing APM does not equal EPM, and how AS differs from a human. We select a replay (named ``PvP\_30'') of AS$^N$ Final. At the time of 4:06, we can see the impressive case, that is although the player's instant APM is twice AS's (461:221), its EPM is a quarter AS's (48:221) (we place the screenshot in Appx.~\ref{appen:neglect in AS}). That is how the agent's actions differing from a human. Its actions are mostly useful actions, and it benefits from the condition when its APM equals humans. Table~\ref{tab:epm and apm} also shows the average value of all AS$^N_P$ replays. Hence, we argue EPM should be a more fair metric to be used.

% Under that case, the agent has more meaningful actions than the humans and can do more things than humans. Hence, the agent has more chance to win. 

% HINT: can be reduced
We formulate $\mathbb{SC}$ which uses an EPM smaller than $x$ as $\mathbb{SC}\{E_x\}$. The $x$ should not only be a simple average number in one game. It could be the average of the past 5 seconds, i.e., the agent can issue a max of up to $15$ actions in the last $5$ seconds if it is $\mathbb{SC}\{E_{180}\}$. We encourage more settings such as $\mathbb{SC}\{E_{150}\}$ and $\mathbb{SC}\{E_{120}\}$. Using less EPM may cause more efficient and effective strategies to be learned.

% MAY: show the experiments

\subsection{Fake Camera} \label{sec:Camera}
\begin{figure}[ht]
    \centering
    \includegraphics[width=0.80\columnwidth]{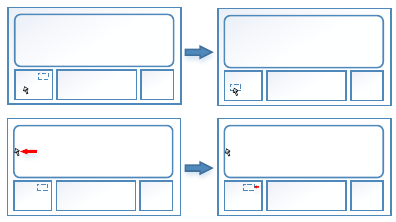}
    \caption{Moving Camera. Top: the clicking way of moving camera. Bottom: the scrolling way of moving camera.}
    \label{fig:Camera}
\end{figure}

\begin{figure*}[t]
    \centering
    \subfloat[]{
        \centering
        \includegraphics[width=0.475\columnwidth]{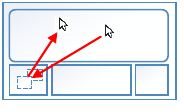}
   }
    \subfloat[]{
        \centering
        \includegraphics[width=0.475\columnwidth]{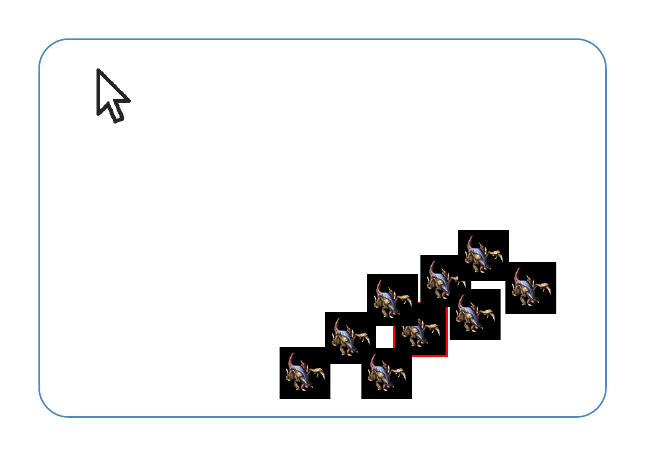}
    }
    \subfloat[]{
        \centering
        \includegraphics[width=0.475\columnwidth]{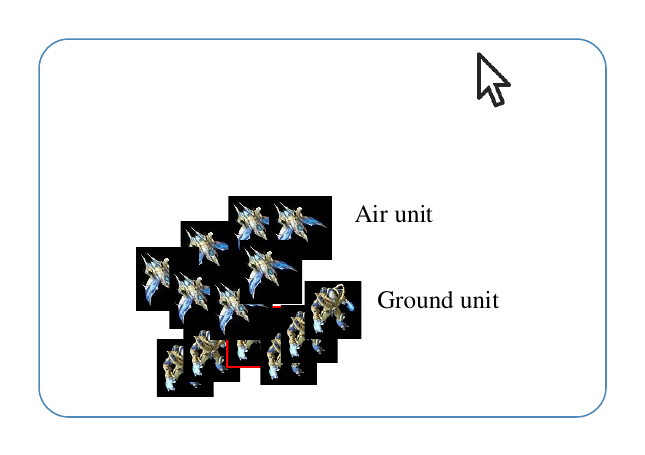}
    }
    \subfloat[]{
        \centering
        \includegraphics[width=0.475\columnwidth]{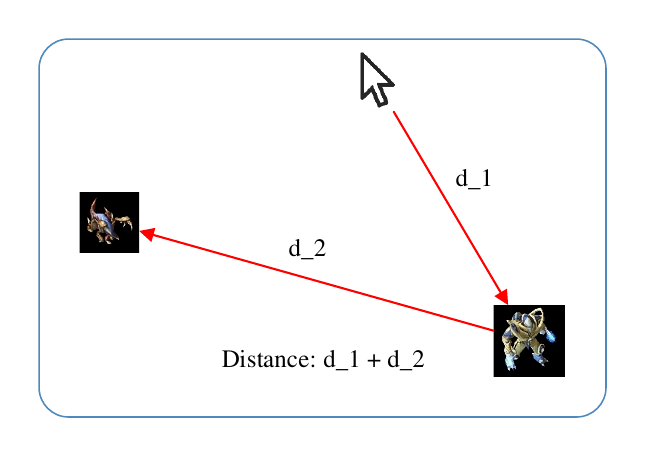}
    }%
    \caption{(a)~Mouse movement trajectory of using clicking way to move the camera. (b)~Zerglings nearly overlap. We want to select the one with a red rectangle. (c)~Air units and ground units overlap. We want to select the ground unit with the red rectangle which is under the air units. (d)~Distance of movement from selecting a unit to giving a target of the action. }
    \label{fig:Control}
\end{figure*}

The camera in SC2 is a window that can slide through the minimap. The area in the camera is the screen. Humans accept two types of visual information, the local one (provided by the screen) and the global one (provided by the minimap). The screen provides most of the information, due to only in it can the player see the health or any other information of the units. In contrast, the minimap provides limited information indicating where the enemy comes or where the mineral locates. Humans can move the mouse in the minimap and do a click to make the camera change to the mouse position where the map locates at. We call this camera moving way ``clicking''. Humans can also make the mouse go towards the edge of the screen, which making the camera move towards the direction orthogonal to the edge, which is called ``scrolling" and is a relatively slow way. The two ones are described in Fig.~\ref{fig:Camera}. The clicking way moves faster but it needs the mouse forth and back between minimap and screen, bearing the overhead of hands moving (see Fig.~\ref{fig:Control}(a)). Hence, some HPs prefer the scrolling way more.

For AS$^B$, DM announced it beats the two professional HPs with an interface having no camera. However, the PS$_2$ at that time only has HA which must use a camera. We suspect that at that time AS$^B$ has already used PS$_3$. By using RA of PS$_3$, AS$^B$ can see the units outside of the camera, and give actions to the units anywhere. So a camera is never needed. This is a huge advantage over humans. DM may realize this and then provided a modified version AS$^B_c$ (meaning AS$^B$ using a camera) to fight against the professional player MaNa (in a Live show). Though they stated AS$^B_c$ has a similar performance as AS$^B$, the result showed a different truth. AS$^B_c$ can't concentrate its attention in the right area of the game. It was deceived by MaNa's confusing units and finally lost the game. 

Actually, there is no support by the BE for using RA with a camera. The reason is that if one uses the RA, all units can be seen and controlled through an entity list as we discussed before. Hence one doesn't need the camera. Under this reasoning, the camera in AS$^B_c$ was actually hand-crafted. The creating way was not known until AS$^N$ is published. 

The camera in AS$^N$ is a 32x20 size rectangle (measured by game units, e.g., a small map has a 64x64 size). It is moved by the ``raw move camera" action in the RA. The camera is actually virtual. The simulating way is to let the opponent units outside the camera have only restrict information (like display type). Note AS can still see and select its own units outside the camera. So this is still different from a true camera. For this reason, we call it the ``fake camera''. This camera causes a problem which is that the AS has advantages over humans. Note humans can do this by using grouping commands (please see SC's wiki), but it is far less flexible than AS's way.

We find AS always uses the clicking way to move the camera (maybe due to AS has no overhead of moving mouse). HPs more often use the scrolling way, which will produce the following effects. Each scroll will generate an action, making the camera action has more proportions of all actions. Hence, AS's non-camera action rate is larger than HP's. E.g., the average of AS$^N_P$ is $0.68$, while HP is $0.50$ (details in  Appx.~\ref{appen:all replay}). Multiplying this with their EPM we can deduce their average ``non-camera EPM'', which are AS ($124.2$) : HP ($78.4$). AS issues $58\%$ more actions than HP per minute, hence have advantages in battle control.

% 
% fake camera make the action space different, a table
% Due to AS has no overhead of moving mouse in ``clicking", AS's most operations are ``clicking". ``clicking" makes AlphaStar has less mouse movement operation. And making the operations 

\subsection{Control Precision}
Here we talk about the control precision problem. Humans may make mistakes while giving commands, especially in a short time. Human actions on SC2 need 3 aspects of control: the precision of action; the precision of selection; and the movement of the mouse. The 1st one means the action should be right. E.g., if an attack action is replaced by a move action, our units may go into a dangerous place (this happens mostly to ranged units). The 2nd one means we should select the right unit. E.g., the unit which has the lowest health should be selected to go back to a safe place in a battle. If we select the wrong one, the injured unit will be destroyed by the enemy. The 3rd one means some action needs the mouse to move to some position. E.g., some action needs first select a unit, then select a target. The mouse should first move from the current position to the unit position and then move to the target position. Such movements need time for humans. But for AS, it is done immediately. In some cases, the 2nd control is hard, see Fig.~\ref{fig:Control}(b) and (c). We show the 3rd case in Fig.~\ref{fig:Control}(d).

% Here we discuss the control precision problem when uses a machine to control actions. 

\begin{figure*}[t]
    \centering
    \includegraphics[width=2.00\columnwidth]{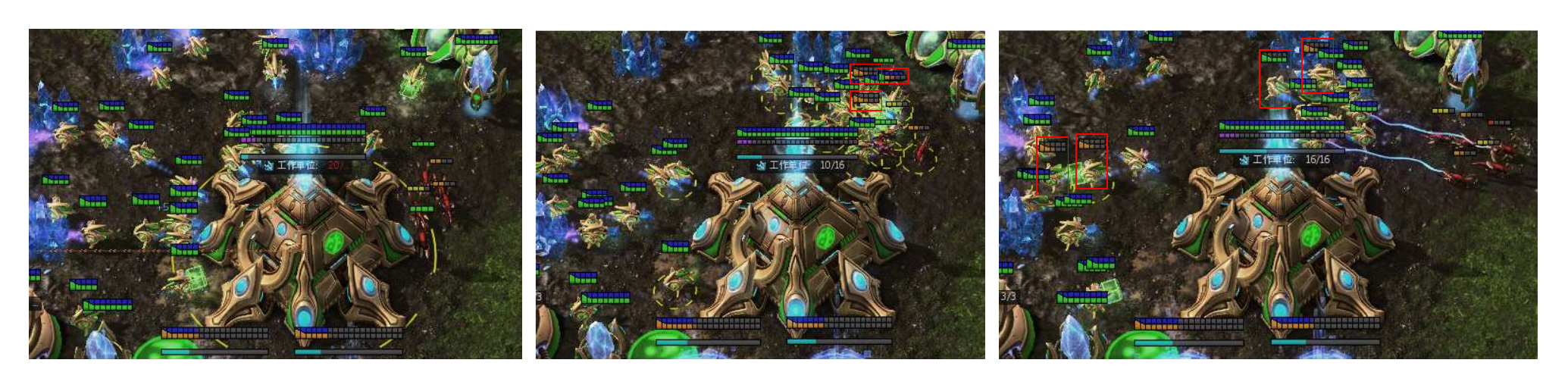}
    \caption{Control precision case. \textbf{Left}: HP's Zerglings want to destroy AS's Probes. \textbf{Middle}: The sensible decision for humans should be letting all the Probes retreat. But, AS's Probes try to defend against the Zerglings, a risky choice. Note Probes and Zerglings overlap. It is difficult to select exactly. Also, injured Probes should go back to the mineral at left. Hence, the mouse needs moving from right to left then back to right, for each action. All need to be done within seconds. \textbf{Right}: However, the result is: Zerglings' attacking fails, no Probe is destroyed. Note red rectangles meaning low health Probes. HP=human player.}
    \label{fig:ZerglingAndProbe}
\end{figure*}

Fig.~\ref{fig:ZerglingAndProbe} shows an extreme case of how AS controls the $3$ aspects which HP is hard to do. We found many of AS's decisions depend on the advantages of the accuracy of control. Assuming AS can't achieve such a perfect defense, it would not choose some overly aggressive tactics to gain advantages. Moreover, after observing most of AS's replays, we found, as long as AS assembles sufficiently large and mixed troops, then it will have almost no disadvantage in the battle. It can accurately control any unit and release skill in a precise location. HP not only needs to select the unit, bearing the risk of selecting the wrong one but also click the position of location afterward, bearing the time overhead of moving mouse. All these put HP in a disadvantaged position. Note AS's accuracy of micro-operation can be seen as a success of RL training. But later we will show AS still lacks a strategic view, and can be defeated by some novel tactics. 

%  AS=AlphaStar.

If we define $\mathbb{SC}$ handled by AS as $\mathbb{SC}\{P_{1.00}\}$ in which $1.00$ indicates the accuracy of control, then we can use $\mathbb{SC}\{P_{0.95}\}$ to express the one we want to face. This $0.05$ error rate can be reflected in any parts of HA and RA. We can use a lower accuracy value during training. E.g., we can train for the $\mathbb{SC}\{P_{0.80}\}$, which may further improve the robustness of the agent.

\subsection{Summary}
Though AS$^N$ has made big progress, we argue its achievements should not be overestimated. The results should be viewed from a more scientific perspective. E.g., the $\mathbb{SC}$ they solve is actually $\mathbb{SC}^r_3\{E_{180}, P_{1.00}\}$, not the one raised in SE. The path between the two may be a long way. Our intention is not to show AS's wrong part, but the neglected part in its presentation. We argue these should be stated clearly. Ignoring them, or explaining in a vague way, will make people overestimate the achievements, adversely affecting the community. From Table~\ref{tab:SC2 problem} we can see $\mathbb{SC}$ solved by AS$^N$ is nearly two generations away from the $\mathbb{SC}$ raised in SE. The AS$^F$ (F refers to Future) is the fictitious version of our imaginary which is the next generation of AS. AS$^F$ uses less human data, gives more constraint on EPM, and has an error rate on control precision. The last problem, proposed here, is the one we think the final SC2 problem is. We argue it may still need decades of years to solve.

% The last $\mathbb{SC}$, proposed here, is the one we think the hardest $\mathbb{SC}$ is. The effective method on that one will have many practical values on real-life tasks.

% The last problem, proposed in our paper, is the one we think the final SC2 problem is. 

\begin{table*}[t]
    \centering
    \scalebox{1.0}{
    \begin{tabular}{l | c c c c }
    \toprule
    Complex Level  & Level-1 & Level-2  & Level-3 & Level-4 \\
    \midrule
    Action Interface  &  $\mathbb{SC}^r$  & $\mathbb{SC}^r$ & $\mathbb{SC}^h$ & $\mathbb{SC}^h$    \\
    Human Data  &  $\mathbb{SC}_3$  & $\mathbb{SC}_2$ & $\mathbb{SC}_1$ & $\mathbb{SC}_0$    \\
    EPM Constraint &  $\mathbb{SC}\{E_{180}\}$  & $\mathbb{SC}\{E_{160}\}$ & $\mathbb{SC}\{E_{140}\}$ & $\mathbb{SC}\{E_{120}\}$    \\
    Camera Usage  &  $\mathbb{SC}\{C_1\}$  & $\mathbb{SC}\{C_1\}$ & $\mathbb{SC}\{C_0\}$ & $\mathbb{SC}\{C_0\}$    \\
    Control Precision &  $\mathbb{SC}\{P_{1.00}\}$  & $\mathbb{SC}\{P_{0.95}\}$ & $\mathbb{SC}\{P_{0.90}\}$ & $\mathbb{SC}\{P_{0.85}\}$    \\
    \midrule
    Problem Def. & $\mathbb{SC}^r_3\{E_{180}, C_1, P_{1.00}\}$ & $\mathbb{SC}^r_2\{E_{160}, C_1, P_{0.95}\}$ & $\mathbb{SC}^h_1\{E_{140}, C_0, P_{0.90}\}$ &  $\mathbb{SC}^h_0\{E_{120}, C_0, P_{0.85}\}$   \\
    Faced by  &  AS$^N$  & AS$^F$ & SE  & Proposed here     \\
    \bottomrule
    \end{tabular}
    }
    \caption{Complex levels of the SC2 problems. AS=AlphaStar. SE=SC2LE. C$_1$=Virtual camera. C$_0$=Real camera.}
    \label{tab:SC2 problem}
\end{table*}

\section{Defects of AlphaStar}
AS achieves GM on BN. However, unlike GM in GO often refer to very professional HP, GM in SC2 could show amateur. E.g, see the HP in the replay of ``007\_TvZ''. We find AS is still far from perfect. We analyze its defects here.

\subsection{Easy To Be Exploited}
\begin{figure}[t]
    \centering
    \includegraphics[width=1.00\columnwidth]{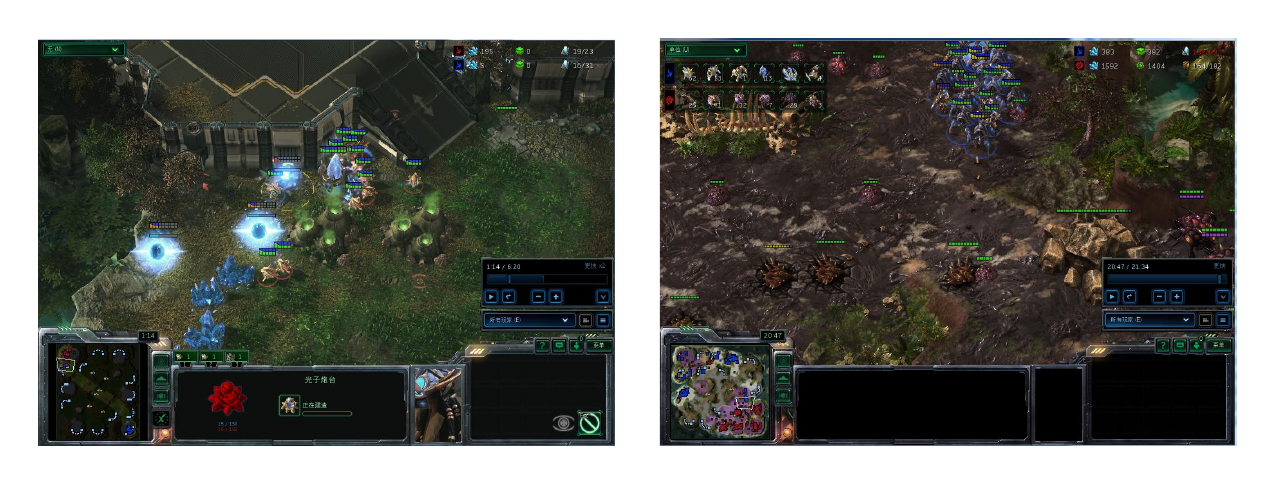}
    \caption{Cases. Left: HP uses Cannon Rush to defeat AS. Right: HP uses Lurkers to beat back AS. HP=human player.}
    \label{fig:cases}
\end{figure}

We first present that AS is still easy to be defeated by novel strategies of HP. In the replay of ``029\_PvP", the HP uses a tactic called ``Cannon Rush" (CR). The key factor of the CR is to secretly build a Pylon near the opponent's base and use the Cannon (CN) to advance gradually to that base. Due to the usage of human data and RL training, AS has learned to defend against this tactic. But the HP changes its strategy a little. He builds two Pylons in a narrow space and one CN inside them. AS uses its workers to attack the Pylons and the CN under construction, but its workers are separated, making its attacking fail to destroy the CN (see left of Fig.~\ref{fig:cases}). When the HP's CN is built, AS loses its advantages and finally be defeated. 

This case shows a defect of the AS. Though if using more RL training days, this problem may be alleviated. We can see that AS still didn't know the key factor to defend against CR. We present more cases in Appx.~\ref{appen:defect of AS}.

% without being seen 

\subsection{Lack of Reasoning}
Here we present a defect of the AS which is that it lacks reasoning. Fig.~\ref{fig:cases}'s right shows a replay of ``028\_PvZ'' where the Protoss (AS) trying to attack the Zerg (HP). The Zerg uses several Lurkers to defend against the AS's attack. Lurkers are invisible units, meaning they can not be seen or attacked until there is a detector (e.g., Observer) to detect them. However, Lurkers can always attack the other units. AS's troops are counter-attacked by the Lurkers and then go back and forth, which have been repeated for several rounds and last for minutes. During this period, AS always doesn't know to produce a detector, though the condition for producing it is already satisfied. This phenomena raise a question, even after 200 years of training (DM states the time of AS training is similar to HP playing for 200 years), does the agent know an invisible unit should be detected by a detector? 

This case raises a thought. For humans, if we know several rules of a game, we can reason and deduce several useful tricks for the game. But for AS which uses induction to learn knowledge, if the knowledge is not shown in human replays, or not be explored in the RL, AS would not know it. For humans, intelligence is composed of induction and deduction. Though the RL equipped with DNN has the power to learn many things, the natural characteristic of it lacks the ability of deduction. We argue that for a complex game like SC which consists of many rules, the introduction of deduction into the AI needs to be considered and more research.

Lurkers are units that would attack, what if an invisible unit which never attacks but stay at AS's home? This case shows in the Live show between AS$^B$ with MaNa. MaNa uses the Observer to collect all AS's information and finally defeats it. AS even knows nothing. AS seems to lack the ability to exploit information, which we would like to see. Unfortunately, we don't see any aspects of it in AS at now.

% in AS$^{B\&N}$ 

% As we discussed before, if AS can win with its superior micro-operations, it doesn't have the need to exploit the power of information. We want to see it trying to hide its information and to gather the opponent's information. This shows cleverness and craftiness and maybe the future directions for AI.

\begin{figure}[t]
    \centering
    \includegraphics[width=1.00\columnwidth]{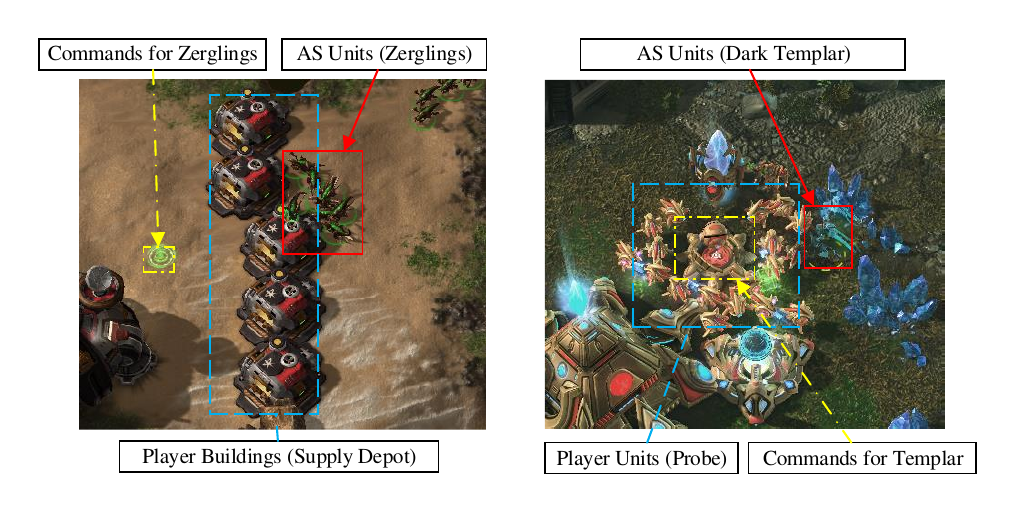}
    \caption{Two Cases. Left: Zerglings want to go in the buildings. Right: Dark Templars want to attack the Cannon.}
    \label{fig:WantToGo}
\end{figure}

A more interesting case occurs due to the lack of reasoning. If HP uses buildings to block its entrance. AS doesn't know how to go in. Its troops will go back and forth around the buildings and waste much time. Fig.~\ref{fig:WantToGo} shows the two cases. Left is Zerglings want to go in the Supply Depots. Right is AS's Dark Templars (DT) want to attack the Cannon (CN), but the CN is surrounded by HP's workers. DT is a melee unit and must first attack the worker can the CN to be attacked. However, AS continuously orders the DT to attack the CN, which makes no effects.

\subsection{Dull Strategy}
We find AS's tactics single and dull. The strategies for each race are nearly the same. E.g., most of tactics of AS$^N_P$ are using Adepts and Oracles to harass, $93.3\%$ tactics of AS$^N_Z$ are first building sub-base, then producing Ravagers to attack. By the league training and a pFSP algorithm, AS learns a policy that is the best response to an un-uniform mixture of policies, which is why AS seems to have one single strategy. However, such behavior is far away from HPs. HPs will try different strategies and have a plan to carry out them over time. The strategies in SC2 should be rich, e.g., Rush, Timing, Economy, Proxy, Fast Technolgy, and so on. However, AS has limited strategies, making its replays boring and lack novelty. Note one reason for creating SC AI is to explore strategies that HPs may never find, which is the interest for the community and the SC players.

% This is different from HP who would try novel tactics and have a plan to carry out the tactics. The tactics of SC should be rich. However, the AS has nearly one single strategy, making its replays boring.

%We think this behavior (imitation of humans) also dues to the z reward in RL which makes AS have a good performance but lost its creativity. Some may say the strategy in AS is the one it finds most effective, hence it doesn't have to change. But even professional HPs would try different strategies, e.g., see GSL 2019 Season 1 final contest Maru vs. Classic. Player Classic uses 6 different tactics in 6 matches.

\subsection{No Planning}
AS uses no planning. In its games, we find it always has only a plan A, no plan B nor plan C exists. That is the most cases when it is defeated by HP. E.g., when its sub-base is destroyed early in the game, it tries to build a new one in a far-away place (making its workers easily being attacked). Also, when its air sneak attack has little effect, it still tries to repeat it. There is no planning module in AS, e.g., MCTS which plays an important role in AG. MCTS provide the way of planning for the future decision. In the current AS, due to the lack of MCTS, the action of $AS^N$ may more consider its current return, instead of planning for the future.

%Due to the lack of MCTS, the action of AS$^N$ more considers its current return instead of future. 

\section{Related Works}
SCC~\cite{Wang2021SCC} was a follow-up work of AS, beating professional HPs in Jun 2020. But, SCC uses the raw interface, and the average APM is $250$-$400$, converting to near $295$ EPM (by $((400-250)/2+250)/1.098=295$, $1.098$ is the ratio of APM on EPM in AS), which is much more than AS$^N$. SCC's paper doesn't mention the camera. From its videos, we infer it uses non-camera. Hence all its EPM are non-camera ones (a large advantage over HPs). Another follow-up work, TStarBot-X~\cite{Han2020TX}, has the pros of open-source. However, it contains some hand-crafted rules, and the performance is also lower than the SCC.

% SCC hence has a great advantage compared to humans. 
% are more likely to 

\section{Future Ways}

\subsection{Using Human Action Space}
AS$^{B\&N}$ all use RA instead of HA, which is not consistent with the problem in SE. We think based on the success of the AS, we should look forward to train in the challenging HA. The hardest part of HA is how to accurately select a unit on the screen by its position coordinate. We think computer vision methods like object detection~\cite{Szegedy2013objectDetection} or segmentation~\cite{Goel2018ObjectSegment} may be candidate solutions. Another untrivial problem of HA is how to train the camera moving. If the camera is exploring randomly at the beginning, the training of the other parts may not be practical. We argue that heuristic ways may be used to control the moving of the camera at the start, after the training of the selection and action is mature, we can concentrate on the training of the camera.

% One way to solve it may be learning macro actions based on HA as proposed in~\cite{pang2019sc2}. 
% Another untrivial part of HA is how to treat the moving camera problem. If the camera is exploring randomly at the beginning of the training, the training of the other part like selection or action may not be practical. We argue that we may use some heuristic ways to control the moving of the camera at the start, after the training of the other part is mature, we can concentrate on the training of the camera.

\subsection{More Fair Comparison with Humans}
We now discuss the fair comparison problem with HP. A fair comparison is hard due to the play of SC consists of 2 parts, the mechanical part, and the intellectual part. AI has advantages in the mechanical part. Yet what we want to test is the intellectual one. One way is to keep the mechanical part as similar as HPs, and then test the other part, as the AS$^N$ does. However, as we analyzed, such a setting has still a long distance from a fair comparison. Some settings may not be adjusted as appropriately as we want. Thus we argue a strict condition should be used, which is that the mechanical part of AI should be weaker than humans. This is aimed at making AI using cleverer tactics, not depending on superior micro-operations. We can also make these settings configurable, thus strength can be adjusted more flexibly and make people select what they want themselves. 

%A more suitable way may be to make these settings configurable. E.g., we can make the parameters of these mechanical settings changed when fighting with humans. Such we can adjust the strength of mechanical part more flexibly and make the people select the appropriate strength they wanted to play with. 

%We can choose a slower or make the precision of action lower. These settings aimed for making AI using cleverer tactics to fight against humans, not depending on its superior micro-operations. 

%A more suitable way may be to make these settings configurable. E.g., we can make the parameters of these mechanical settings changed when fighting with humans. Such we can adjust the strength of mechanical part more flexibly and make the people select the appropriate strength they wanted to play with. 

% E.g., AS$^N$ uses the RA, meaning it doesn't need to have any selection operation like humans; AS$^N$ uses a fake camera, meaning they can select its units anywhere and become more flexible; the action of AS is executed $100\%$ precisely, meaning it has an advantage over humans on accuracy; AS$^N$ doesn't have the need to move the mouse in the screen, meaning its action can be executed much faster than humans. 

\subsection{Fewer Resources and Open Source}
AS used 4,200 CPU cores and 256 TPU cores, plus 44 days for training an agent. We think training an agent which has comparative performance with much fewer resources should be explored, like in~\cite{Liu2021ThoughtGame}. Also because the SC is the simulation of a battle or war, the learning abilities of the agent may also facilitate the research on the wargame or controlling the combat unit in the real world, thus the research by closed-source may exacerbate the military imbalance in the world. The research by open-sourced may be the way to remedy the balance.

% \subsection{Fewer Resources}
% AS uses about 4,200 CPU cores and 256 TPU cores for training one agent which is a huge cost. AS also has a training day of 44. We think in the future we should explore training a similar agent which has comparative performance with much fewer computing resources and time. Some works(~\cite{Liu2021ThoughtGame}) have tried these, but their performance is only tested with the built-in AI. Training a computer that has a similar level as a master-level player on one single commercial server in one day may be a potential direction for future research.

% \subsection{Open Source}
% We argue that an important direction for AI on SC2 should be open-sourced. As we analyzed in the SC2 problems section, due to the setting of imperfect information and larger state and action space, the SC2 problem shows more similarities as the real-life problem and thus has more practical value. Also because the SC is the simulation of a battle or war, the learning abilities of the agent may also facilitate the research on the wargame or controlling the real combat unit in the real world, thus the research by closed-source may exacerbate the military imbalance in the world. The research by open-sourced should be the way to maintain this balance.

\section{Conclusion}
In this paper, we present a critical rethinking of AS. Through experiments and analysis, we show that the problem AS solves is actually a sub-problem of the original one. By its replays, we show its defect and analyze the reason connecting to its architecture. Finally, we give directions for further research on SC2 problems.
 
\newpage
\bibliographystyle{aaai}
\bibliography{liu}

\newpage

\appendix

%Here we present the features of the SC2 problem, more cases for neglect in AS, more cases for defects of AS, and an analysis of all AS Replays.

\begin{appendices}

\section{More Details of SC2} \label{appen:more of SC2}
%Since SC2 is currently a very difficult environment for AI, a brief introduction is given in this section. 

In SC2, players can choose one from four 4 settings (Protoss, Terran, Zerg, Random) to play. The Random setting is to randomly select one from 3 races. The 3 races (Protoss, Terran, Zerg) have almost completely different types of units, but a strong balance. There can be 3 or more players in one game, but we research here is mostly 2 players (a Zero-Sum game). Within the 2 players, there is a host player to create a room. Then a participant player can go into the room. The host player can select a map from a map list that contains more than 10 maps. The 2 players select their races. Then the host can start the game. Though the participant can not decide which map to use or when to start the game, he can exit the room at any time. When playing in BattleNet, the map is selected by the system, and the probability the two players encounter is based on their ranked level. Several features of SC2 make it a suitable environment for testing general AI.

% At the start of the game, the 2 players are born in different locations on the game map (usually at the opposite corners). Players need to win the game by destroying all the opponent's buildings (note, destroying opponent's all soldiers is not a necessary option). 

\textbf{1. Imperfect information}.
%\subsection{Imperfect information}
Unlike Go, players in SC2 cannot see the opponent's information. Except for all the visible areas of our units, the rest of the game map is shrouded in a layer of fog called fog-of-war. There are two layers of this fog. The first layer means the areas that we have never explored, which is black fog. Under this black fog, none of the information can be seen. When our units have explored this area once, this layer is replaced by a second layer of fog called gray fog. Under the gray fog, player can only see information such as terrain or the last buildings he saw. The existence of the fog-of-war requires players to carry out appropriate scouting and exploration and also increases the requirements for game-theoretic tricks.

\textbf{2. Various types of units}
%\subsection{Various types of units}
In Go, each stone has the same meaning. On the contrary, in SC2, each unit type has unique characteristics and is different from others, which is similar to a real war. E.g., Terran has the Marine, Tank, Battlecruiser, and so on. These types share similarities with the real ones in the world. Although Protoss is a fantasy race, its units have historical shadows. E.g., Zealot corresponds to swordsmen in history, and Templar corresponds to wizards. There are counter relationships between these unit types, which form a ``Rock-Paper-Scissors" cycle, so there is no absolutely invincible unit. In addition, combine with the considering of imperfect information, the game-theoretic tricks in SC2 need to be very skillful. E.g., player 1 (P1) detects that player 2 (P2) has produced unit A, so P1 produces many units B that can counter A. But P1 did not expect P2's unit A to be a guise. P2 deliberately lets P1 see the production of unit A and actually produces unit C which can counter unit B. So in the final battle P2 wins. This kind of game technique is not uncommon in SC2 games. 

\textbf{3. Speed of Operations}. 
%\subsection{Operation and response speed}
In terms of action space, SC2's is quite large (see Table~\ref{tab:game versions} to see a comparison with other games). It is hard to learn and choose one action in such a tremendous action space. Moreover, unlike Go, SC2 also requires players to have a fast reaction and operation speed. E.g., some actions should be done in hundreds of milliseconds (on the contrary, one action in Go may be done in dozens of seconds). Because of this difference, some think SC2 is more difficult than Go due to the speed requirements places high requirements on the processing speed of AI. Others believe that AI gains more advantages by its operation and response speed over humans. That is, the AI can do actions in milliseconds, which means it can do more action than humans. Due to the micro-operations in SC2 have a large influence on the battle results, some AI bots that can use much more actions in a certain time (a very high APM) have gained advantages, e.g., a bot like Automaton 2000 Micro. Although AS claims that it did not rely on these micro-operations to gain advantages, we found that the victory of many games of AS is established by the accumulation of these small micro-operation advantages, not the overall strategic view like humans. E.g., AS can beat some HPs (human players) on the frontal battlefield but be defeated by other HP's novel tactics. 

\begin{table}[t]
    \centering
    \scalebox{0.9}{
    \begin{tabular}{l | c}
    \toprule
    Game  & Action Space \\
    \midrule
    MountainCar &  2  \\
    Gridworld & 4  \\
    Atari & 17  \\
    Go & 19 x 19  \\
    SC2 & 564 x 128 x 2 x 512 x 512 x 256 x 256  \\
    \bottomrule
    \end{tabular}
    }
    \caption{Comparison of action spaces in different games. SC2 has 564 raw actions. Delay is 1 to 128. Queued is 0 or 1. Entity list length is 512, so spaces for selected units and the target unit are 512. The world size is used as the target position, which is 256 x 256. }
    \label{tab:game versions}
\end{table}

Though SC2 is a suitable environment for AI, it has some disadvantages. E.g., SC2 is a heavy game. A Windows version of SC2 needs about 10G disk space, meaning installing SC2 is not as easy as Go. Running an SC2 process needs hundreds of megabytes of memory, meaning it is hard to run hundreds of processes in parallel in a normal commercial server. There are multi-versions of SC2. The version of it has developed from 3.16.1 to 5.0.0. Different versions have dissimilarities and are not compatible which makes training and testing inconvenient. Comparing to the simulating speed of Go, SC2 is also much slower. One episode of SC2 may need minutes to simlulate, even using the fastest speed.

% (e.g., step\_mul is set to 8)

\clearpage
\newpage

\section{Terminology of SC2} \label{appen:term of SC2}
Due to SC2 is relatively complex, some players will use some terms to express some matters briefly and clearly. We present some common terms here, for reference.

Compared to Go, SC2 is much harder for humans to get start. People can learn the rules of Go in several minutes and begin to play (though it may spend dozens of years to be professional). Ont the contray, the rules of SC2 need several hours learning. SC2 is complex, having a bunch of rules, hundreds of unit types, and dozens of maps. For a human player, he may need several days to learn to defeat the normal built-in AI, and  a long time for learning before he can play against other human players. 

For readers of interest in SC2 AI, we recommend referring to SC2 wikis to know knowledges of SC2 (some famous wiki is like liquipedia and so on). We present here some terms using in our paper.

\textbf{Rush}: A tactic, which is getting the advantages in the early game by attacking as fast as possible. Sometimes a good Rush will win the game.

\textbf{Proxy}: A tactic, called in Chinese as "Ye". It can be used with Barrack and Gateway, like "Proxy Barrack" or "Proxy Gateway". These means to build these buildings in the field near the opponent's home. It is key to not be scouted by the opponent. Then we can gain the most advantages when launch attacking (such as Rush).

\textbf{Timing Attack}: A tactic, which mostly refers to an attack at a suitable time, such as when the opponent is building its sub-base (time when the opponent having the least defend power). If the timing is right, such attack will make the opponent lost advantages, so we could win.

\begin{figure}[h]
    \centering
    \includegraphics[width=1.00\columnwidth]{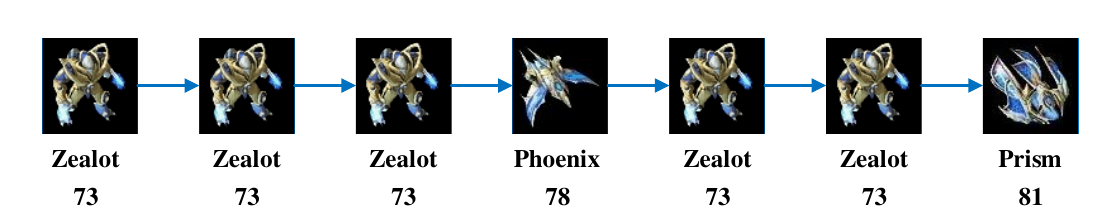}
    \caption{A case of build order. It can also includes building. The unit type IDs for Zealot, Phoenix, and Warp Prism are 73, 78, 81 respectively. }
    \label{fig:bo}
\end{figure}

\textbf{Build Order}: A term, which refers to an order in which all units are built. The units here include units and buildings. It can be regarded as a list. Whenever a new building or producing command is completed, the unit will be appended to the list. Fig.~\ref{fig:bo} is an illustration. In AS it is a list as $\{73, 73, 73, 78, 73, 73, 81, \dots\}$. Since the units in the list are related in order, the Levenstein distance is used in AS to calculate its reward.

% \{Zealot, Zealot, Zealot, Phoenix, Zealot, Zealot, Warp Prism, ...\}

% Zealot 73
% Phoenix 78
% Warp Prism 81

\textbf{Unit Counts}: A term, which is also named as ``Build Units''. This refers to the units currently the player owns. The units here include units and buildings. It can be regarded as a set. The method of ``bag of words'' can be used in getting the Unit Counts. This statistic value shows the composition of our or enemy's units. Since the order is irrelevant, Hamming distance is used in AS to calculate the relevant rewards.

\clearpage
\newpage

\section{More Information of AS} \label{appen:more of AS}
We now discuss the disadvantage and advantages of AI compared to humans on SC2. The disadvantage is that the AI may not be smart enough. E.g., script-based AI makes its responses not robust. However, the AI which is trained by RL may have the ability to respond effectively in various scenarios. On the contrary, the advantages of AI are many: 1. AI can obtain accurate information about the units, but humans need to observe them with the eye or remember them; 2. AI can respond very quickly, e.g., it can perform an action at each step, making the highest APM can go to $22.4 * 60=1344$ (in real-time mode, SC2 advances 22.4 steps within one second), which exceeds the limit of humans; 3. AI's accuracy of AI is high, and it does no wrong actions, opposed to humans; 4. AI has no time overhead to move the mouse, which makes its reaction speed much faster than humans.

Although SC2 is a real-time game, it is also running according to the game step. Only after all players (including the built-in AI) have selected the action in this step, the environment will advance to the next step (this is similar to Go). The speed at which the environment advances affects the effect of the game display. In the real-time mode, the environment advances at a speed of 22.4 FPS. That is, 22.4 steps are advanced every second, so the time of each step is about 44.6 milliseconds (denoted as T$_r$). It can be seen that this response time requirement is much greater than that of Go. During training, the real-time mode can be closed, making if the agent's inference time is larger than T$_r$, the environment will wait for the agent to give an action. At this time, from the perspective of the player, the game screen image is paused, similar to a turn-based game. When playing against people, the environment should enable the real-time mode, which is that, if the agent doesn't give a response within T$_r$, this step will be skipped. Because of this feature, AI on SC2 is hard to use some time-consuming technologies, such as MCTS. When using neural networks for decisions, the network's inference speed is also important.

In SC2LE, there are restrictions on the information that AI can gather. However, in AS$^B$, these restrictions are lifted. E.g., it can get complete control over the information of each unit, instead of inferring through the feature maps before. Although these advantages are controlled to a certain extent in AS$^N$, AI still has several advantages compared to that when the SC2LE was proposed. Meanwhile, the advantages of AI are not limited (such as control accuracy). This makes humans face unfair situations when fighting against AI.

\begin{table}[h]
    \centering
    \scalebox{1.0}{
    \begin{tabular}{l | c c c c }
    \toprule
    Type  & AS$^Z$ & AS$^B$  & AS$^B_c$ & AS$^N$ \\
    \midrule
    Raw Units  & No & Yes & Yes & Yes  \\
    Action Space  &  Human  & Raw & Raw & Raw    \\
    Camera  &  Yes  & No & Fake & Fake   \\
    Race  &  Terran  & Portoss & Portoss & All   \\
    APM  &   $\approx 500$  & $\approx 500$ & $\approx 500$ & $\approx 200$    \\
    \bottomrule
    \end{tabular}
    }
    \caption{AS versions. AS$^Z$ refers to the baseline in SC2LE. Note raw units are the Raw Observation (RO).}
    \label{tab:AS versions}
\end{table}

Since our paper is an analysis of AS, some sources of information are stated here. AS contains 2 published versions, one is the Blog version, denoted as AS$^B$, and the other is the Nature version, denoted as AS$^N$. The differences announced by DeepMind between the two are as follows: 1. AS$^N$ places more restrictions on APM than AS$^B$; 2. AS$^N$ uses a camera to restrict the information gained, compared to AS$^B$; 3. AS$^N$ trained 1 agent for each race, a total of 3 agents, while AS$^B$ only trained 1 agent for the Protoss race. The game result of 10:1 (AS against 2 professional human players) uses the AS$^B$ version. AS$^N$ contains three versions: Supervised, Mid, and Final. The replays provided by AS$^N$ consists of 90 replays for the Supervised version, 180 replays for the Mid version, and 90 replays for the Final version. AS$^N$ provides supplementary materials, which are pseudo-codes (including 4 files) and 1 detailed architecture. 

We show the characteristics and differences of each version of AS, please refer to Table~\ref{tab:AS versions}. We also present all the versions of AS$^N$ in Table~\ref{tab:AS Nature versions} in which the  AS$^N_P$ Final is the strongest agent, which we mostly discuss in our paper.

\begin{table}[h]
    \centering
    \scalebox{1.0}{
    \begin{tabular}{l | c c c }
    \toprule
    Type  & AS$^N_T$ & AS$^N_Z$  & AS$^N_P$ \\
    \midrule
    Supervised  & - & - & - \\
    Mid  &  -  & - & -  \\
    Final  &  4.9.3  & 4.9.3 & 4.10.0 \\
    \bottomrule
    \end{tabular}
    }
    \caption{Versions of AS$^N$. The number in the cell is the SC2 versions of these replays of that version. Actually, AS$^N_P$ uses the latest SC2 version, which may mean it has more training time compared to the agents of the other two races.}
    \label{tab:AS Nature versions}
\end{table}

\clearpage
\newpage

\section{Experiments of RA vs. HA} \label{appen:experi of ravsha}
\begin{figure}[t]
    \centering
    \subfloat[]{
        \centering
        \includegraphics[width=0.475\columnwidth]{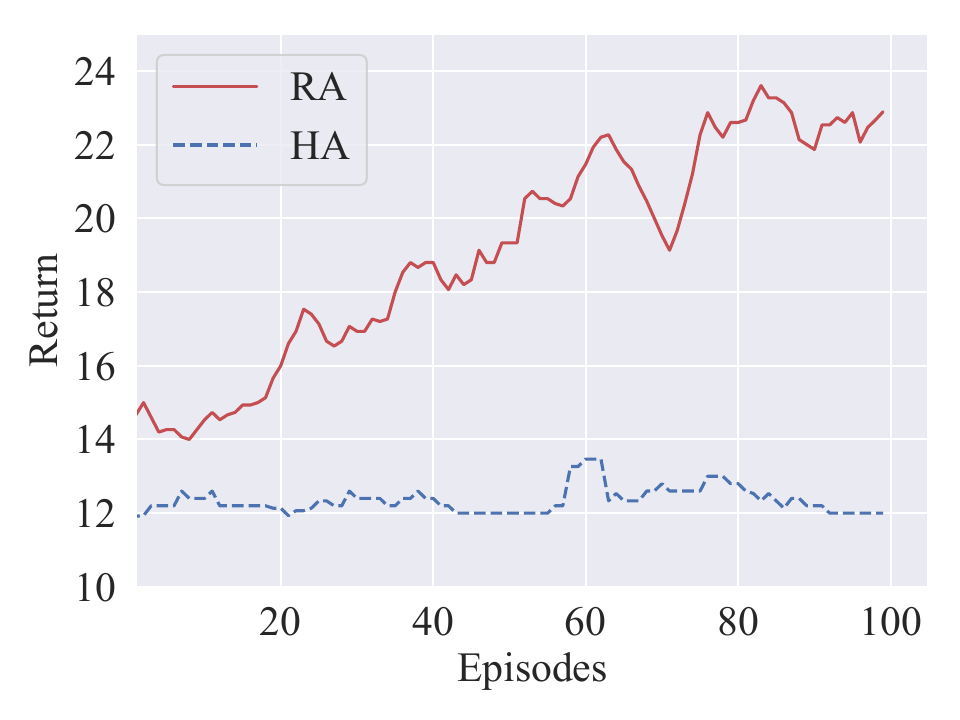}
   }
    \subfloat[]{
        \centering
        \includegraphics[width=0.475\columnwidth]{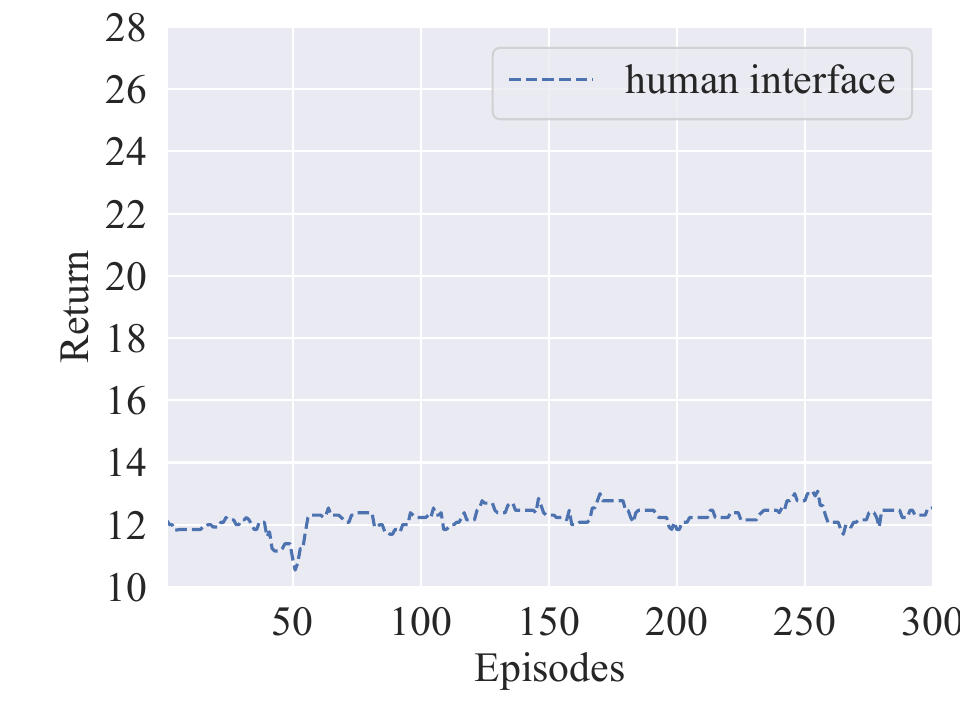}
    }%
    \caption{(a)~Training curves on the two action spaces for 100 episodes. The results are averaged by 3 runs. (b)~Training curve on the HA for 300 episodes. RA=raw actions. HA=human actions.}
    \label{fig:rawvshuman4}
\end{figure}

Here we show the experiments of comparing training on the raw actions (RA) with on human actions (HA). These experiments are based on a prototype problem in SC2, which is to control the Protoss race to set up its economy. The more workers, the more minerals per minute the agent will get. So the agent's goal is to produce (or called train in SC2) workers as many as possible in a maximum of 4500 game steps. The agent has two action spaces, human actions, and raw actions. The definition of the two action spaces is in the main body of the paper, but here we do a little simplification for training speed. E.g., the RA has 3 actions (build Pylon, train Probe, and call Probe to collect mineral). Initially, we have 12 workers (Probes). The main goal is to train Probes, and the reward is from the difference in the number of workers between the two adjacent states (state of now and previous state). However, the food supply has a maximum of 15, if we do nothing, we can only train a max of 15 workers. So we must also build some Pylons to increase the supply. The entity list contains our workers and the Nexus (base) which could train workers. If the agent wants to train a Probe, it should provide the right index in the entity list to select the Nexus in the selected units head. Or, if the agent wants to build a Pylon, it should provide the index of the workers in the entity list. Thus, the agent should give the right commands by which action to do (in the action type head) and which unit to do this action (in the selected unit head). This is not easy compared to vanilla RL training which only needs to give one action type. But due to the max number of the entity list in our experiment is set to not high (8), the time for exploration should be enough to get a good policy. Before inputting to do processing, we make the entities in the entity list have an order. That is, the first one (index 0) is the Nexus, and the rest are Probes sorted by their tag values (the first Probe has the index of 1, the last Probe has the index of 7). This is to make the processing before and after the model have the same order. The Probes having an index larger than 7 are ignored. The red solid line in Fig.~\ref{fig:rawvshuman4}(a) is the training curve of RA. It can be seen that the return (cumulative rewards in one episode) can rise quickly. 

Compared to RA, HA has 5 actions, which are ``build Pylon'', ``produce Probe'', ``collect minerals'', ``select point'', and ``move camera''. The two more actions than RA are used for selecting units. E.g., if the agent wants to build a Pylon, it should first select a Probe. If the agent wants to produce Probes, it also should first select the Nexus (base). Thus, the right action should consist of two actions in continuous two steps (select and do actions). Note that the target position (the coordinate) of the select action is based on the content in the screen, so a ``move camera'' is needed in the action space. The training results are shown as the blue dashed line in Fig.~\ref{fig:rawvshuman4}(a). The training on HA seems hard. The learning curve is hard to rise. To test if providing more training time, the results whether can be improved, we change the overall training steps to 3 times as the previous one. But in Fig.~\ref{fig:rawvshuman4}(b), we can see that even giving more training time, the learning curve still can not rise, illustrating the training difficulties on the HA. 

We conducted our experiments on a Linux server that has an Intel(R) Xeon(R) Gold 6248 CPU 2.50GHz with 48 cores, 500GB memory, 1T disk space, and 8 NVIDIA Tesla V100 32G (we only use one of them). We use the SC2 version of 4.0.3 and use PyTorch 1.5 as the deep learning framework. We use the PySC2$_3$ version as the interface. We write our experiment code base on an open-source reproduction of the AlphaStar, the mini-AlphaStar, which is a mini-scale reproduction of the AS. In our simplification, we remove unuseful feature inputs in the entity encoder, spatial encoder, and scalar encoder, and cut the queue head, delay head, and target unit head from the architecture for simplification. We use the food of workers as the reward without any scaling.

We use Adam as our optimization algorithm. We use the learning rate as 1e-3 because we are training from scratch for the RL agent. Other hyper-parameters of Adam are beta1=0, beta2=0.99, epsilon=1e-5, weight\_decay=1e-5. We use V-trace as our training algorithm (we found adding the UPGO algorithm doesn't change the performance by much). The batch\_size is set to 128. We also use an entropy loss of which weight is set to 3. The other training details are presented in the code appendix.

\begin{figure}[h]
    \centering
    \subfloat[]{
        \centering
        \includegraphics[width=0.475\columnwidth]{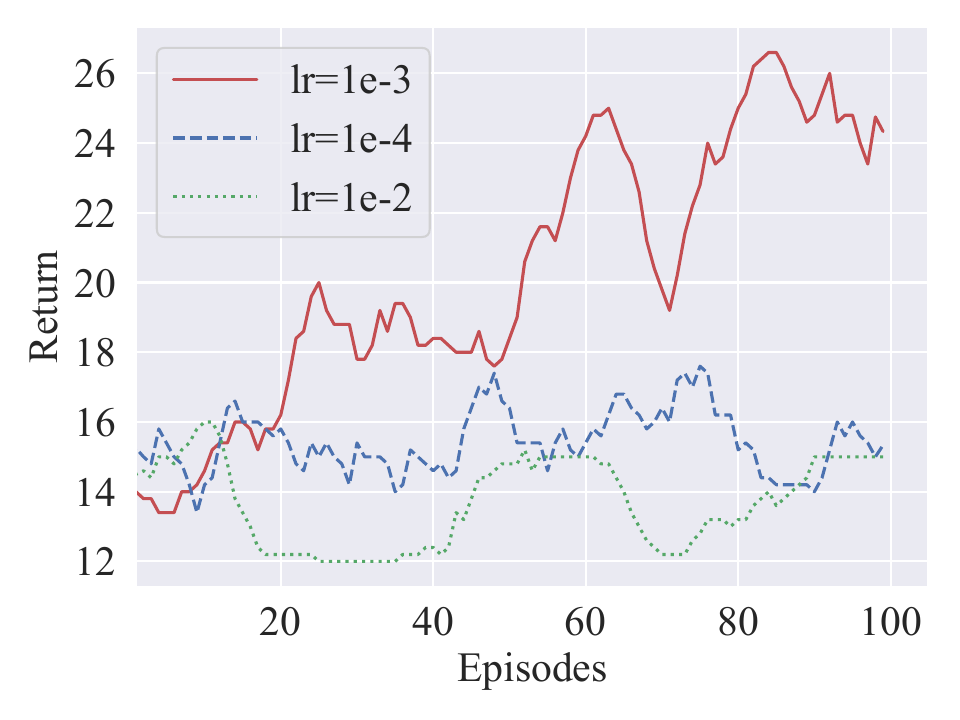}
   }
    \subfloat[]{
        \centering
        \includegraphics[width=0.475\columnwidth]{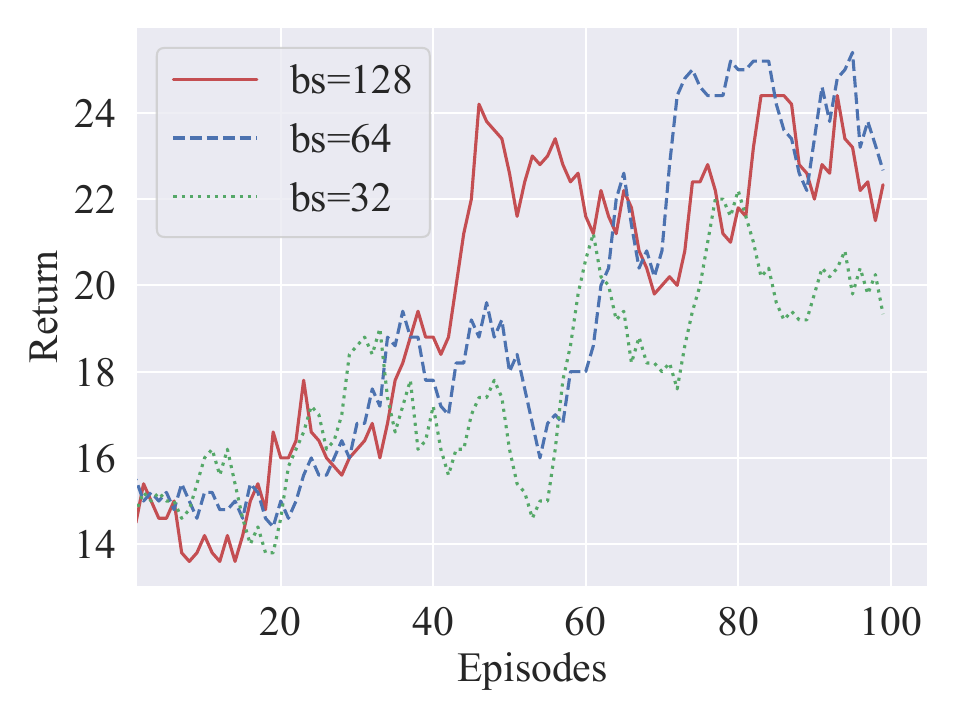}
    }%
    \caption{(a)~Comparison of different learning rates (lr) on RA. (b)~Comparison of different batch sizes (bs) on RA. }
    \label{fig:hyperparameters}
\end{figure}

We use grid search to decide our hyper-parameters. E.g., we choose the learning rate by running the experiment three times, using 1e-2, 1e-3, and 1e-4 respectively. As shown in Fig.~\ref{fig:hyperparameters}(a), we select the learning rate as 1e-3 due to it performs best. We also choose the value of the batch size in this way, as shown in Fig.~\ref{fig:hyperparameters}(b).

\clearpage
\newpage

\section{Neglect in AS (More Cases)} \label{appen:neglect in AS}
Here we show more cases of neglect in AlphaStar.

\subsection{Impact of APM and EPM}
\begin{figure}[h]
    \centering
    \includegraphics[width=1.00\columnwidth]{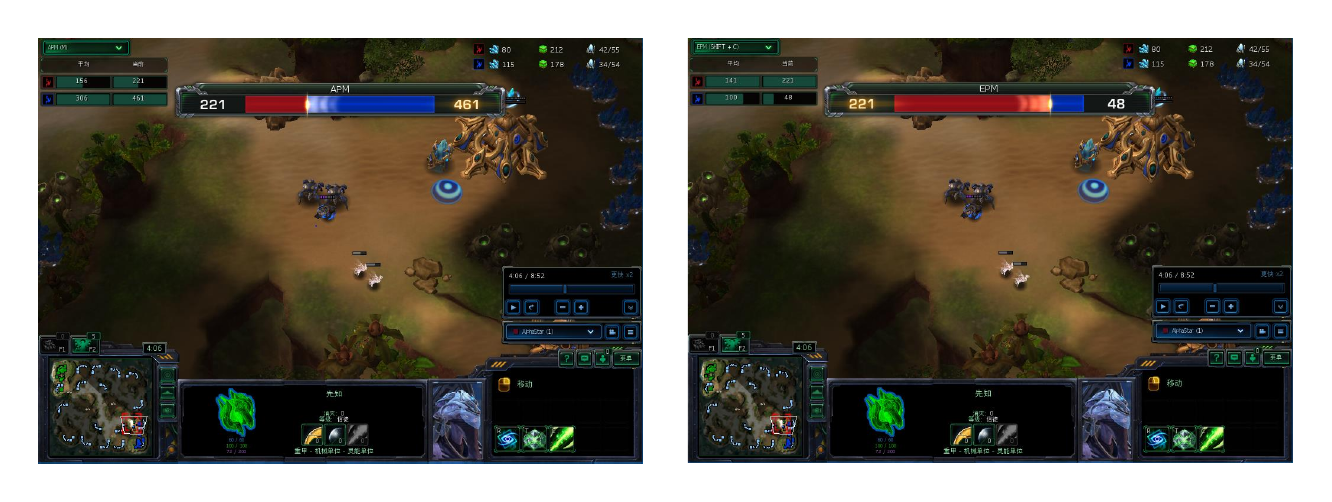}
    \caption{The case of the APM (left) versus EPM (right) at the same time. The red bar is AS's, and the blue one is the HP's. Note not always the AS's EPM equals its APM, which is a special case here.}
    \label{fig:apmvsepm}
\end{figure}

In Fig.~\ref{fig:apmvsepm} we show the screenshot of the special case, which is that AS's instant APM is near half of HP (human player)'s (221:461), but AS's EPM is 4 times HP's (221:48). This impressive case shows that how many actions in a human player's APM are unuseful actions, compared to an AI's.

\subsection{Fake Camera}
Here we show the difference in the proportion of camera actions caused by the difference in the way that AS and HP use the camera, please see Table~\ref{tab:camera rate}. 

\begin{table}[h]
    \centering
    \scalebox{1.0}{
    \begin{tabular}{l | c c c }
    \toprule
    Type  & camera\_op & all\_op  & no\_camera\_rate  \\
    \midrule
    AS  &   322.79 &	969.55 &	\textbf{0.6823} \\ 
    HP &	859.13 &	1747.41 &	0.5088 \\
    \bottomrule
    \end{tabular}
    }
    \caption{Average (all AS$^N_P$ replays) for the count of camera operations, all operations, rate of none-camera operations. AS=AlphaStar. HP=human player. Bolds are better values.}
    \label{tab:camera rate}
\end{table}

It can be seen that AS's no\_camera\_rate is $0.6823$, which is larger than HP's $0.5088$. This proportion can be multiplied with EPM, to get the ``effective non-camera actions'' in one's actions.

\subsection{Control Precision}
\begin{figure}[t]
    \centering
    \includegraphics[width=1.00\columnwidth]{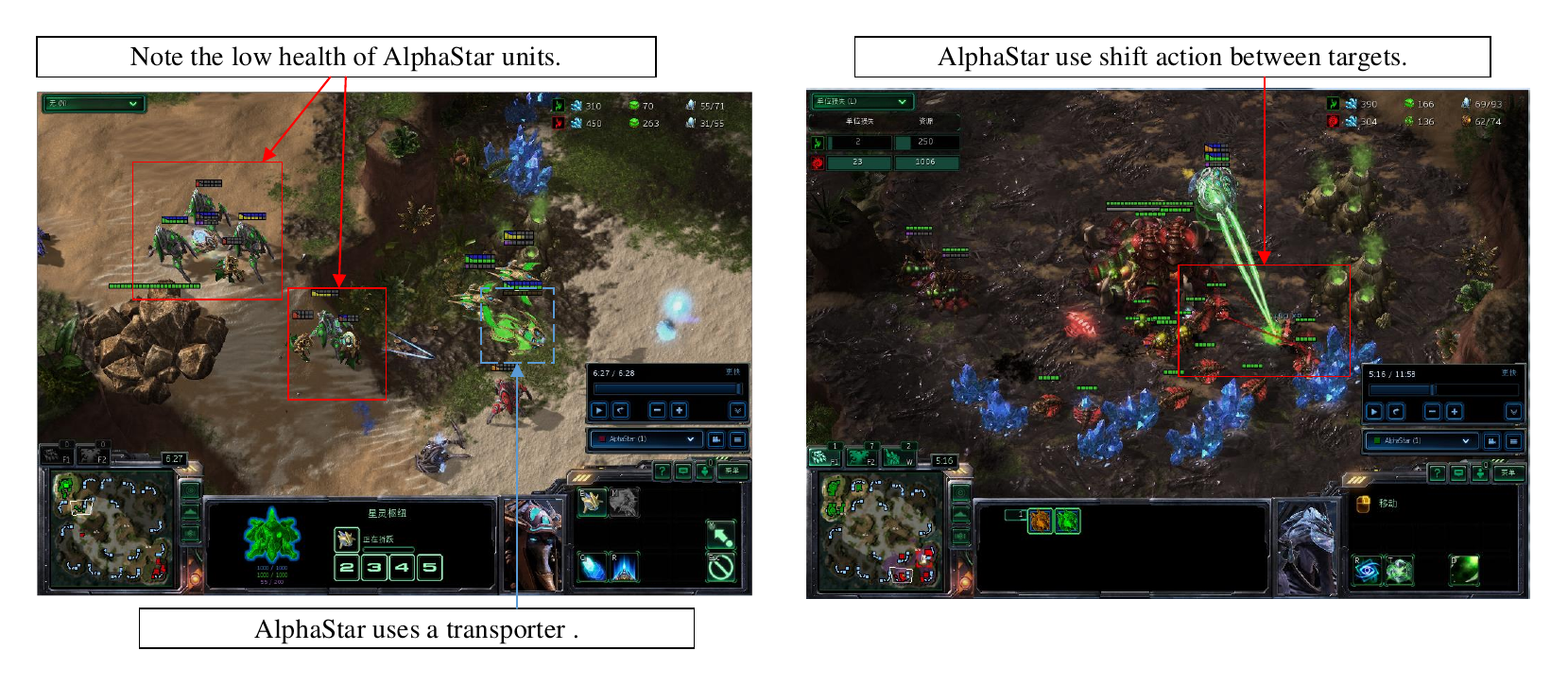}
    \caption{Two examples about control accuracy. Left: A result of one battle between AS and HP. Right: AS's Oracle is harassing the opponent's workers. AS=AlphaStar. HP=human player.}
    \label{fig:ControlAccuarcy}
\end{figure}

We give more cases of AS's control precision here and use Fig.~\ref{fig:ControlAccuarcy} as an example. The left in the figure shows a result of one battle. The green units are soldiers of AS and the red units are ones of HP. Displayed above the heads of these units are their health points. After the battle, HP's units were nearly wiped out, in contrast AS preserved most of its soldiers. Actually, at the beginning of the battle, the numbers of soldiers on both sides were nearly the same. AS had made most of its units survived by virtue of its accurate control. It can be seen that most of the AS's soldiers have low health points, showing the careful protection of AS on its soldiers. Within the battle, AS often used the transport aircraft (called Warp Prism in SC2) to load the unit which is being attacked, and then unload it in a safe place. By this control, AS can defeat troops that are more than its. The right in the figure shows that when AS used the Oracle to do a sneak attack, it used the shift operation to put all the target workers in the queue for an instant. Due to the precise operation of AS, it can complete this process in milliseconds, while HPs need to bear the possibility of errors of selection or overhead of mouse movement.

\begin{figure*}[t]
    \centering
    \includegraphics[width=2.00\columnwidth]{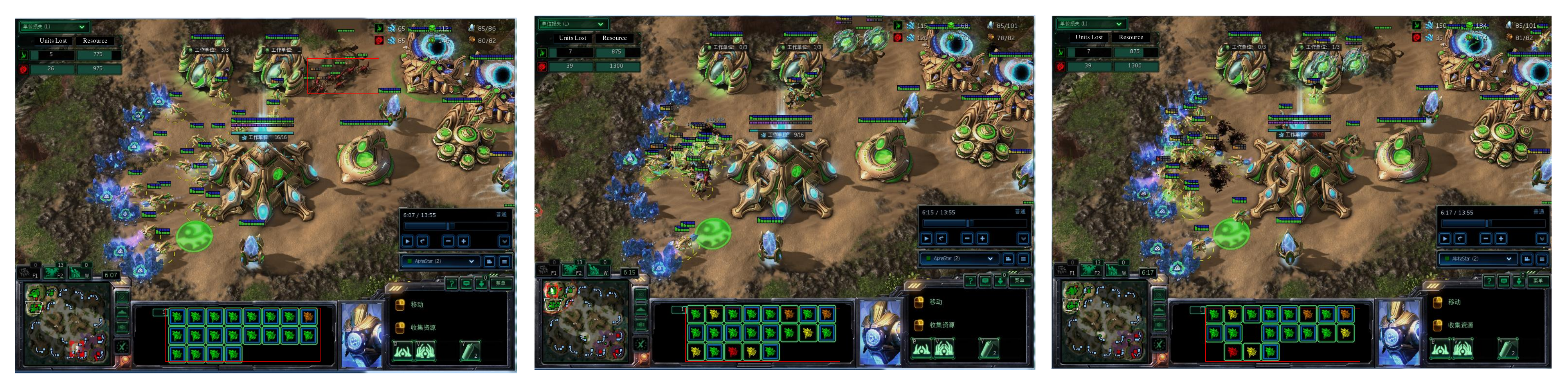}
    \caption{\textbf{Left}: HP's Zerglings go in the AS's base and want to destroy AS's Probes. \textbf{Middle}: AS uses its Probes to defend against the Zerglings, which is a risky choice for humans. \textbf{Right}: However, after the battle, the Zerglings are all wiped out. And the Probes have only lost 2 of them. The red rectangle is the health of all Probes. Note that many Probes have low health, but only two ones are destroyed after the battle.}
    \label{fig:ZerglingCannotKill2}
\end{figure*}

Here we give another example of AS using Probes to resist Zerglings attacking, which is in Fig.~\ref{fig:ZerglingCannotKill2}. There are more Zerglings here than that in the example in the main paper, but AS only paid the price of 2 Probes to defend successfully and eliminate all the attacking Zerglings, which is very fully seeing the ability of AS's micro-operation (hard for HPs).

\clearpage
\newpage

\section{Defects of AS (More Cases)} \label{appen:defect of AS}
Here we show more cases of defects of the AlphaStar.

\subsection{Easy To Be Exploited}
\begin{figure}[h]
    \centering
    \includegraphics[width=1.00\columnwidth]{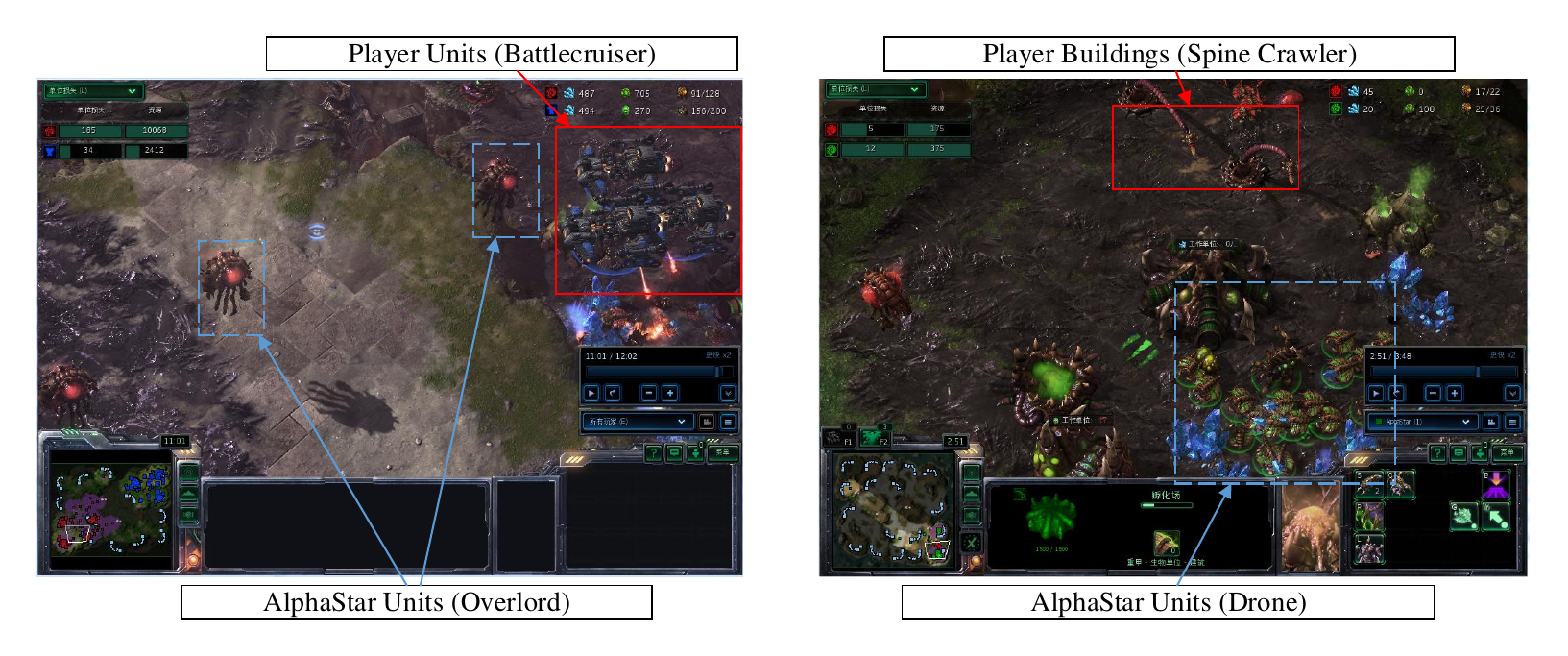}
    \caption{Two examples of being exploited by novel tactics. Left: AS is defeated by HP's Battlecruisers' sneak attack. Right: AS is defeated by HP's Crawler Rush. AS=AlphaStar. HP=human player.}
    \label{fig:NovelTactics}
\end{figure}

There are two examples of how AS was defeated by human player's (HP) new tactics, which is shown in Fig.~\ref{fig:NovelTactics}. The figure on the left shows an example. In the early stage, AS had an advantage and was about to win. However, HP used Battlecruiser to continuously harass AS through the skill of Battlecruiser: the Tactical Jump. The Battlecruiser killed many AS's workers and Overlords, making the economy of AS suffered a great loss. When HP built more and more Battlecruisers, the original AS's advantages disappeared. In the end, AS didn't even know how to play the game, making its Overlords go directly to the touch Battlecruiser (this is undoubtedly a suicidal behavior because Overlord has no offensive ability). It can be seen from this replay that AS is not good at adapting to the opponent's new tactic. The figure on the right shows that AS was being attacked by HP's Spine Crawler Rush. When the race of AS is Zerg, we find it rarely used its workers to scout. Therefore, once it is targeted by the opponent's rush tactics, it is easy to be defeated. 

%When AS$^N_Z$ is playing, it rarely scouts. In the ZvZ game, one of the Zerg players can choose Crawler Rush. Due to AS$^N_Z$ does not scout, it was unable to deal with such tactics.

\subsection{Lack of Reasoning}
\begin{figure*}[t]
    \centering
    \includegraphics[width=2.00\columnwidth]{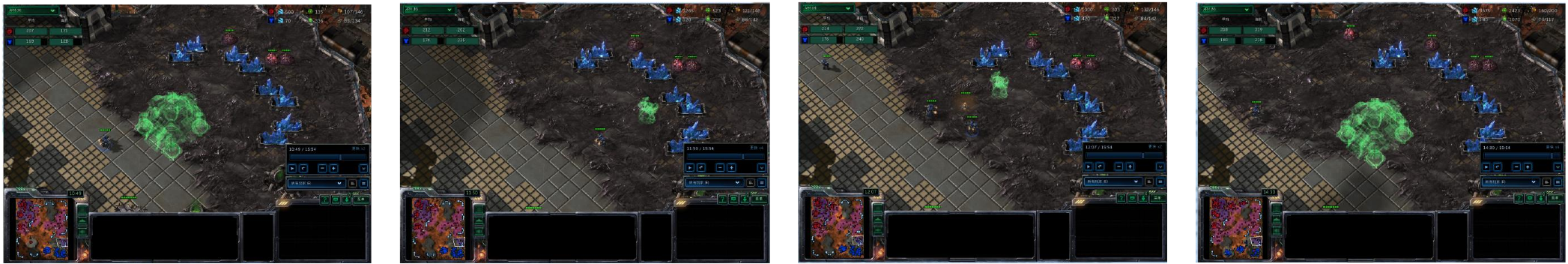}
    \caption{(1) AS wanted to build a base (Command Center) on the Creep at 10:49. (2) AS wanted to build a Missile Turret on the Creep at 11:50. (3) AS secondly wanted to build a Missile Turret on the Creep at 12:07. (4) AS wanted to build a Command Center on the Creep at 14:30.}
    \label{fig:WantToBuild}
\end{figure*}

Fig.~\ref{fig:WantToBuild} shows that an example that AS doesn't learn one knowledge of the game. At the beginning of the game, HP laid a Creep on the 2nd mining area of the AS (which should be the position to build the AS's sub-base). AS's race is Terran. Due to the rules of SC2, Terran buildings can not be built on the Creep. But, AS didn't learn this knowledge. From the 10:00 of the game, it had always wanted to build buildings there. But the game engine keeps returning the hint which is that the building can't be built here. This process had been repeated for several rounds and last for 5 minutes. Obviously, these unuseful actions made AS waste much time and economy.

\subsection{Dull Strategy}
The strategies of the AS for each Race are nearly the same (see Table~\ref{tab:AS Protoss 1}, Table~\ref{tab:AS Terran 1}, and Table~\ref{tab:AS Zerg 1}). The ``Strategy Order'' column shows the strategy of the AS agent at that replay. E.g., the start strategies of AS$^N_T$ are mostly producing a Reaper, then transferring to Hellions, and then using Banshees to conduct aerial harassment. The start strategies of AS$^N_Z$ are mostly building a sub-base quickly (which is called the Economy tactic in SC2, a risky strategy), then transferring to use Roach. AS$^N_P$ will build two Adepts to harass the enemy's workers and then use Oracles to conduct harassment, of which process is repeated in all matches, even against all races. 

By the league training, AS has learned a policy that is the best response to an un-uniform mixture of strategies. That is the reason why AS seems to have only one single strategy. However, such behavior is far away from humans. HPs will try different strategies and have plans to carry out the strategies over time. E.g., see GSL 2019 Season 1 final contest Maru vs. Classic. Player Classic uses 6 different tactics in 6 matches. However, the AS's strategies are fewer, making its replays boring and lack novelty. 

\begin{figure}[t]
    \centering
    \includegraphics[width=0.90\columnwidth]{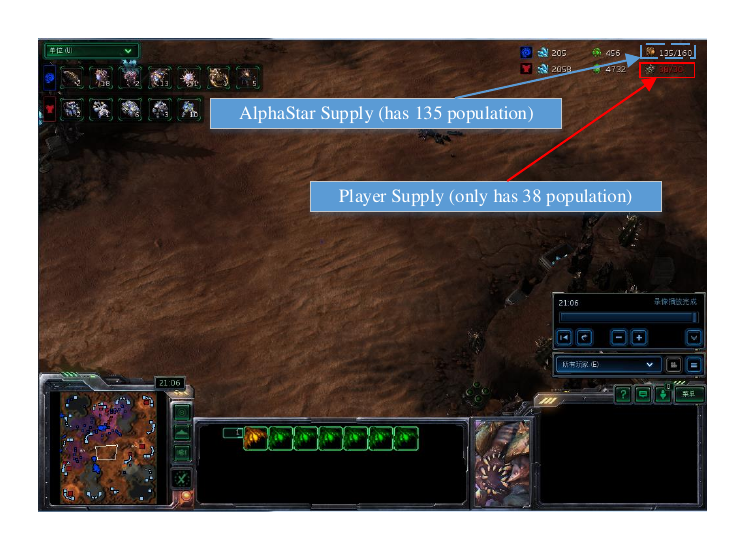}
    \caption{One case about swapping home. Note AS has more population (number of units) than HP, but its buildings are being destroyed faster than HP. HP=human player.}
    \label{fig:SwitchHome}
\end{figure}

\subsection{No Planning}
We now show an interesting phenomenon about ``swapping home'' (or called ``Base Trade'' in SC2). In the replay of Fig.~\ref{fig:SwitchHome}, AS had advantages in the latter part of the game. But HP sent troops to attack the AS's base. Meanwhile, AS's troops were also attacking HP's base. We call this situation ``swapping home". Just like we introduced the SC2 game before, the victory condition of the game is to destroy all the opponent's buildings. Therefore, who destroys buildings first wins. AS had more units, but it obviously didn't understand this rule. Hence, it was at a disadvantage in this situation. HP used fewer soldiers, but he destroyed AS's buildings faster. At the same time, HP was very careful to save its own buildings and prevent AS from finding them, while AS's soldiers were still wandering around on the map. AS neither tried its best to find the opponent's building nor did it build its own building in a hidden place. So it finally lost. From this point, we can see that AS still hadn't learned the victory rule of SC2, and it didn't have the ability to plan in such a situation, thus losing the game.

\clearpage
\newpage

\clearpage
\newpage

\section{Scouting in AS} \label{appen:scout in AS}
\begin{figure}[h]
    \centering
    \includegraphics[width=0.95\columnwidth]{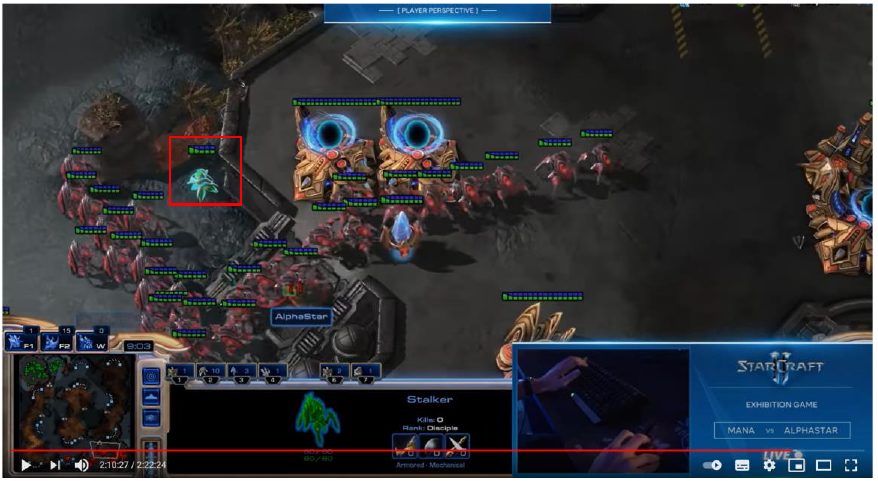}
    \caption{The Live show between AS$^B_c$ and MaNa. This picture is in AS's base. The one in the red rectangle is MaNa's Observer.}
    \label{fig:liveshow}
\end{figure}

In the ``Lack of Reasoning'' subsection, we show AS is counter-attacked by an HP's Lurkers. The Lurkers are invisible units that would attack, meaning that the information ``we are under attack by unknown ones so we need some detectors to see them" could be learned through training. What if the invisible units could not attack? If these units do nothing, but only stay at the AS's base, AS may be unaware of it. This case arose in the match between AS$^B_c$ and MaNa in the Live show (see Fig.~\ref{fig:liveshow}). The human player MaNa produced an Observer, a Protoss detector unit that is also invisible. MaNa ordered the Observer to stay at AS's base, seeing near everything of it (like how many troops, the technology tree, and etc). The AS's information is nearly open up to MaNa, but AS doesn't know that. Finally, MaNa used this information to plan a strategy and defeated AS. 

This is an interesting case which arises several thoughts. Does AS know it has already been seen everything by others? Does it know the power of information and have any ideas to hide its information? Does it try to deceive its opponent to gain advantages? Does it try to gather information about its opponent to build an anti-strategy? As we discussed in the APM and control precision subsection, if AS can win with its superior micro-operations and no-mistake actions, it wouldn't have the need to exploit the power of information. We want to see it trying to hide its information, trying to gather the opponent's information, and trying to deceive the opponent. This shows cleverness and craftiness and maybe the future directions for AI. Unfortunately, in this version of AS, we didn't see any aspects of it.

AS still lacks attention to the importance of scouting. In some replays, it doesn't send a worker to scout (which is opposed to what humans would do). In all AS$^N_Z$ replays, it always uses the Overlord to scout instead of using a worker to do scouting. Though this is a common way but the Overlord is much slower than a worker, making some information may be missing. Actually. some human Zerg players will use the worker to do scouting. Due to this reason, the AS$^N_Z$ was defeated by the opponent which used some Rush tactics in some cases. In AS$^N_T$ replays, it mostly doesn't do scouting. Only if its opponent race is Terran, it will use the worker to patrol around its base to find the opponent's Proxy Barracks. We call this ``Proxy Scouting". The AS$^N_P$ is the agent which used scouting most often. But in some matches, it still didn't scout. 

Nevertheless, AS's scouting performance is better than previous scripted bots, and it really uses its units to collect some information (like what AS$^N_T$ and AS$^N_P$ do). The decision of not scouting in some cases may be due to the results of RL training. E.g., it wants to maximize economic income, so it does not want to waste a worker to do scouting. Of course, in many cases, it has paid the price (being beaten by the HPs' Rush). For the scouting information, see the ``Scout'' column in Table~\ref{tab:AS Protoss 1},~\ref{tab:AS Terran 1}, and~\ref{tab:AS Zerg 1}.

\clearpage
\newpage

\section{Description of All Replays} \label{appen:all replay}
We give the content (including some information we gather by programs) of all replays of AS Final here, see Table~\ref{tab:AS Protoss 1},~\ref{tab:AS Terran 1}, and~\ref{tab:AS Zerg 1}. The description of replays shows a brief description of game process in Table~\ref{tab:AS Protoss 2},~\ref{tab:AS Terran 2}, and~\ref{tab:AS Zerg 2}. 

\begin{table*}[t]
    \centering
    \scalebox{0.8}{
    \begin{tabular}{c | c | c | c | l | c | c c | c c | c c c | c c c}
    \toprule
    ID &Type & OL &Res. &Strategy Order &Scout &E$^A$   &E$^P$	&A$^A$	&A$^P$  &CO$^A$   &AO$^A$	&NCR$^A$	&CO$^P$   &AO$^P$   &NCR$^P$ \\
    \midrule
    01 & PvZ & GM & Win   &SubBase$\rightarrow$Adept$\rightarrow$Oracle   &has   &180	&195	&191	&298&311&925&0.664 &1346&2222&0.394 \\
    02 & PvZ & GM & Win   &SubBase$\rightarrow$Adept$\rightarrow$Oracle   &has   &190	&173	&205	&260&440&1322&0.667 &1263&2411&0.476 \\
    03 & PvZ & GM & Win   &SubBase$\rightarrow$Adept$\rightarrow$Oracle   &has   &198	&183	&215	&251&551&1482&0.628 &1553&2991&0.481 \\
    04 & PvZ & M  & Win   &SubBase$\rightarrow$Adept$\rightarrow$Oracle   &has   &183	&244	&198	&391&334&1010&0.669 &651&1554&0.581 \\
    05 & PvT & M  & Win   &SubBase$\rightarrow$Adept$\rightarrow$Oracle   &has   &170	&135	&188	&326&192&624&0.692 &539&1045&0.484 \\
    06 & PvT & M  & Win   &SubBase$\rightarrow$Adept$\rightarrow$Oracle   &has   &169	&103	&190	&134&164&551&0.702 &447&902&0.504 \\
    07 & PvT & M  & Win   &SubBase$\rightarrow$Adept$\rightarrow$Oracle   &has   &187	&156	&208	&237&307&913&0.664 &578&1427&0.595 \\
    08 & PvZ & GM & Los  &SubBase$\rightarrow$Adept$\rightarrow$Oracle    &has   &173	&135	&202	&345&167&473&0.647 &600&993&0.396 \\
    09 & PvZ & GM & Win   &SubBase$\rightarrow$Adept$\rightarrow$Oracle   &has   &184	&89	    &204	&242&101&444&0.773 &292&511&0.429 \\
    10 & PvP & M  & Win   &Adept$\rightarrow$Oracle                       &none  &153	&104	&170	&156&237&683&0.653 &557&944&0.410 \\
    11 & PvT & M  & Win   &Adept$\rightarrow$Oracle                       &has   &202	&141	&223	&272&429&1292&0.668 &588&1556&0.622 \\
    12 & PvZ & M  & Win   &none                                           &-     &-	    &-	    &-	    &-  &-  &-  &-  &-  &-  &-\\
    13 & PvZ & M  & Win   &SubBase$\rightarrow$Adept$\rightarrow$Oracle   &has   &184	&163	&201	&321&356&1091&0.674 &665&1625&0.591 \\
    14 & PvZ & GM & Win   &SubBase$\rightarrow$Adept$\rightarrow$Oracle   &has   &212	&190	&226	&266&269&933&0.712 &847&1908&0.556 \\
    15 & PvP & GM & Win   &Adept$\rightarrow$Oracle                       &none  &181	&124	&197	&145&294&864&0.660 &672&1355&0.504 \\
    16 & PvP & M  & Win   &Adept$\rightarrow$Oracle                       &none  &170	&96	    &186	&118&234&811&0.711 &1016&1524&0.333 \\
    17 & PvP & GM & Win   &Adept$\rightarrow$Oracle                       &none  &177	&158	&197	&226&168&649&0.741 &564&1201&0.530 \\
    18 & PvP & M  & Win   &Adept$\rightarrow$Oracle                       &none  &153	&153	&167	&202&273&844&0.677 &690&1532&0.550 \\
    19 & PvP & M  & Win   &Adept$\rightarrow$Oracle                       &none  &187	&162	&204	&279&234&873&0.732 &609&1462&0.583 \\
    20 & PvP & GM & Win   &Adept$\rightarrow$Oracle                       &none  &166	&149	&181	&225&208&670&0.690 &476&1016&0.531 \\
    21 & PvZ & GM & Win   &SubBase$\rightarrow$Adept$\rightarrow$Oracle   &has   &207	&208	&225	&284&655&1732&0.622 &1682&3479&0.517 \\
    22 & PvP & GM & Win   &Adept$\rightarrow$Oracle                       &none  &181	&152	&201	&223&388&1038&0.626 &776&1660&0.533 \\
    23 & PvZ & GM & Win   &SubBase$\rightarrow$Adept$\rightarrow$Oracle   &has   &177	&173	&194	&249&427&1172&0.636 &972&2047&0.525 \\
    24 & PvP & GM & Los  &Not seen                                        &none  &186	&111	&202	&145&42&296&0.858 &127&347&0.634 \\
    25 & PvP & UN & Los  &Adept$\rightarrow$Oracle                        &none  &183	&166	&204	&289&405&1190&0.660 &1047&2158&0.515 \\
    26 & PvP & M  & Win   &Adept$\rightarrow$Oracle                       &none  &188	&144	&205	&253&499&1398&0.643 &1179&2357&0.500 \\
    27 & PvP & M  & Win   &Adept$\rightarrow$Oracle                       &none  &160	&144	&180	&207&208&650&0.680 &866&1547&0.440 \\
    28 & PvZ & M  & Los  &SubBase$\rightarrow$Adept$\rightarrow$Oracle     &has   &219	&261	&236	&348&984&2553&0.615 &3019&6231&0.515 \\
    29 & PvP & GM & Los  &Not seen                                        &-    &198	&132	&226	&208&147&723&0.797 &611&1214&0.497 \\
    30 & PvP & M  & Win   &Adept$\rightarrow$Oracle                       &none  &163	&127	&179	&271&337&911&0.630 &683&1456&0.531 \\
    \midrule
    - &avg            & -      & -   &-                                  &-    &182	&154	&200	&247&322 &969 &0.682 &859 &1747 &0.509\\
    \bottomrule
    \end{tabular}
    }
    \caption{AS$^N_P$ replays. Res.=AS's result. OL=opponent level. GM=Grand-Master. M=Master. UN=Unranked. E$^A$=EPM for AS. E$^P$=EPM for player. A$^A$=APM for AS. A$^P$=APM for player. CO=camera op. AO=all op. NCR=non-camera rate.  }
    \label{tab:AS Protoss 1}
\end{table*}

% E$^A$=EPM for AS. E$^P$=EPM for HP. A$^A$=APM for AS. A$^P$=APM for HP. CO=camera op. AO=all op. NCR=non-camera rate.

\begin{table*}[t]
    \centering
    \scalebox{0.8}{
    \begin{tabular}{| c | l |}
    \toprule
    ID & Description of the Game Process \\
    \midrule
    01  &  At first, GM is harassed by AS's Adepts and lost many workers. FB: GM has 30 people less than AS and loses.                                \\
    02  & GM has died many workers by AS at first. FB: AS has a population of 200 to beat GM which has 150 \\ 
    03  & AS shows using workers to kill Zerglings. FB: GM uses Mutalisk to fight for AS and then lost. But if GM chooses to swap homes, he may win.          \\
    04  & PL resists Adepts well. But AS's 3 Oracles forcibly attack (face the defensive tower), kill many workers. FB: PL died many troops by Disruptor.          \\
    05  & PL resists 1st wave of Oracles well. But the 2nd defense failed and many workers died. FB: PL's troops were divided by Force Field and defeated. \\
    06  & PL resists Adepts not bad, but can not resist the Oracles. PL's Hellion harassment was also wiped out by AS's Oracles. FB: PL loses.            \\
    07  & In the early stage, many PL's workers were eliminated by AS's Adepts and Oracles. FB: All PL's troops were destroyed by AS's troops.                \\
    08  &  AS's Probe detected that the Zerg (PL) had not established a sub-base, but AS did not react to it. AS was defeated by Crawler Rush.    \\
    09  &  In the first three minutes, PL's 10 workers were killed by AS's harassment. PL's mentality has been hit hard. PL directly quit the game.     \\
    10  & PL took strict precautions but 20 workers were still killed by suicide harassment. PL attacks and AS builds 3 Battery. PL loses the battle.       \\
    11  & AS's 3 Adepts killed PL's many workers. PL produced no Tanks. FB: Many PL's troops died due to the precise micro-operation of AS's Disruptor.                     \\
    12  & One second after the game starts, the player quits the game.                                 \\
    13  & AS's harassment at the beginning of the game killed 20 to 30 PL's workers. FB: AS's army has a population of 150, and it won.                  \\
    14  & GM resists harassment well. GM used Nydus to carry out a sneak attack but did not harm AS's economy. GM lost after several waves of battles.        \\
    15  & GM defended well first. But one accidentally, 2 Oracles of AS rushed in and killed 8 workers. Then 6 workers were also killed by AS's 4 Adepts.             \\
    16  & AS's 3 Adepts and 1 Oracle killed 20 workers. FB: PL can't beat AS.               \\
    17  & AS kept killing PL workers with 2 Adepts. Sometimes it not only kills but also retreats successfully. FB: AS uses Prism to save many units.        \\
    18  & AS used two Adepts and Oracle to kill more than a dozen workers of PL. FB: AS wiped out all PL troops.               \\
    19  & PL adopted Stalker Rush in the early stage. It worked very well at first, but the situation changed after AS produced Immortal and Prism.          \\
    20  &  PL did not perform well against AS's Adepts offense and Oracle offense. PL's economy was destroyed.         \\
    21  &  AS uses a mixed troops containing Disruptor and Warp Prism to gain a huge advantage to beat the PL.        \\
    22  &  AS and PL have the same start building order. However, operations of AS are more refined and has an advantage in the exchange of battle units.  \\
    23  &  AS uses mixed troops, and incredible micro-operations which humans hard to do, to win.            \\
    24  &  PL uses Cannon Rush. PL blocked the road with 3 Pylons and built Cannon inside. The workers of AS were wiped out.                   \\
    25  &  All the preventive details of PL are done to the extreme. AS does not make advantages over PL. On the contrary, AS was suppressed. FB: PL won.           \\
    26  & The details of PL's Cannon Rush were not done well and be defend by AS. After that, PL failed to resist the AS's harassment and was defeated.                   \\
    27  & In the beginning, AS's Adepts harassed PL. PL attacked the base of the AS but troops were wiped out due to AS's micro-operations. PL quits.           \\
    28  & In the beginning, AS was sneak attacked by Zerg using Nydus. Zerg then produced Lurker. After its Observer was killed, AS did not replenish one.                \\
    29  & AS was suppressed by PL's Cannon Rush. AS defended with 1 Immortal and 2 Battery. But 2 Oracles of the PL killed all AS's workers. PL won.          \\
    30  & At first, AS used 2 Adepts to harass. Then AS used 2 Oracles to harass (AS used the shift operation very quickly). PL did not harass. PL lost.  \\
    \bottomrule
    \end{tabular}
    }
    \caption{Description of AS$^N_P$ replays. FB=Final Battle. PL=player.}
    \label{tab:AS Protoss 2}
\end{table*}

% AS showed frequent two-wire operations. 
\begin{table*}[t]
    \centering
    \scalebox{0.8}{
    \begin{tabular}{c | c | c | c | l | c | c c | c c | c c c | c c c}
    \toprule
    ID &Type & OL &Res. &Strategy Order &Scout &E$^A$   &E$^P$	&A$^A$	&A$^P$  &CO$^A$   &AO$^A$	&NCR$^A$	&CO$^P$   &AO$^P$   &NCR$^P$ \\
    \midrule
    01&TvT& GM & Los &SubBase                        &proxy &160	&125	&188	&190&182&585&0.689 &405&980&0.587 \\
    02&TvT& GM & Win &SubBase$-$Hellion$-$Liberator  &none  &149	&97	    &160	&152&142&449&0.684 &408&722&0.435 \\
    03&TvP& GM & Los &Bunker$-$SubBase$-$Banshee     &none  &170	&162	&188	&204&727&1794&0.595 &2231&4227&0.472 \\
    04&TvZ& GM & Los &SubBase$-$Banshee              &none  &179	&236	&197	&389&656&1576&0.584 &1360&3807&0.643 \\
    05&TvT& M  & Win & Bunker$-$SubBase$-$Banshee    &proxy &155	&155	&174	&191&261&681&0.617 &861&1659&0.481 \\
    06&TvZ& M  & Win & SubBase$-$Hellion             &none  &177	&174	&193	&366&357&949&0.624 &815&1920&0.576 \\
    07&TvZ& GM & Win &SubBase$-$Hellion              &none  &201	&195	&214	&348&386&1141&0.662 &937&2178&0.570 \\
    08&TvT& UN & Los &SubBase$-$Banshee              &proxy &187	&242	&202	&349&546&1501&0.636 &1686&3948&0.573 \\
    09&TvZ& GM & Los &SubBase$-$Hellion              &none  &204	&201	&228	&282&706&1934&0.635 &2345&4351&0.461 \\
    10&TvZ& M  & Win & SubBase$-$Hellion             &none  &186	&219	&201	&319&434&1203&0.639 &890&2657&0.665 \\
    11&TvP& M  & Win & SubBase$-$Hellion$-$Banshee   &none  &197	&181	&214	&256&589&1461&0.597 &1399&2834&0.506 \\
    12&TvP& M  & Win & SubBase$-$Hellion             &none  &131	&132	&145	&261&108&347&0.689 &319&727&0.561 \\
    13&TvP& GM & Los &SubBase$-$Hellion$-$Liberator  &none  &174	&142	&195	&218&348&1031&0.662 &1077&2052&0.475 \\
    14&TvZ& GM & Win &SubBase$-$Hellion              &none  &176	&183	&191	&331&414&1020&0.594 &967&2137&0.547 \\
    15&TvT& M  & Win & Bunker$-$Tank$-$Banshee       &proxy &144	&155	&165	&265&106&357&0.703 &402&962&0.582 \\
    16&TvP& GM & Los &SubBase$-$Hellion$-$Banshee    &none  &195	&202	&215	&338&672&1694&0.603 &2094&4097&0.489 \\
    17&TvZ& M  & Win & Bunker$-$Marine$-$Tank        &none  &191	&188	&209	&333&426&1097&0.612 &1284&2403&0.466 \\
    18&TvZ& GM & Win &SubBase$-$Hellion              &none  &193	&250	&204	&428&324&1078&0.699 &722&1745&0.586 \\
    19&TvZ& M  & Win & SubBase$-$Hellion             &none  &193	&181	&212	&268&352&1052&0.665 &909&2032&0.553 \\
    20&TvZ& M  & Los & SubBase$-$Hellion             &none  &160	&156	&177	&215&558&1407&0.603 &1833&3346&0.452 \\
    21&TvP& M  & Win & SubBase$-$Hellion$-$Banshee   &none  &146	&130	&160	&289&253&651&0.611 &536&1261&0.575 \\
    22&TvT& GM & Win &SubBase$-$Hellion              &none  &172	&195	&187	&337&292&820&0.644 &700&1684&0.584 \\
    23&TvZ& M  & Win & SubBase$-$Hellion             &none  &198	&141	&211	&223&296&867&0.659 &733&1391&0.473 \\
    24&TvP& GM & Los &SubBase$-$Banshee              &none  &180	&174	&195	&272&490&1126&0.565 &1578&2714&0.419 \\
    25&TvZ& M  & Win & SubBase$-$Hellion             &none  &191	&195	&205	&333&381&1099&0.653 &946&2037&0.536 \\
    26&TvZ& GM & Los &SubBase$-$Hellion              &none  &201	&191	&219	&319&864&2144&0.597 &2012&4153&0.516 \\
    27&TvP& M  & Win & SubBase$-$Hellion$-$Banshee   &none  &124	&90	    &139	&159&130&366&0.645 &366&626&0.415 \\
    28&TvZ& M  & Win & SubBase$-$Hellion$-$Banshee   &none  &177	&253	&194	&349&317&869&0.635 &704&1733&0.594 \\
    29&TvT& GM & Los &Bunker$-$SubBase$-$Hellion     &proxy &173	&180	&190	&331&327&999&0.673 &813&1863&0.564 \\
    30&TvZ& GM & Los &SubBase$-$Hellion              &none  &204	&256	&222	&445&695&1659&0.581 &1792&3594&0.501 \\
    \midrule
    -&Avg& - & - &- &- &176 &179 &193 &292 &411&1098&0.635 &1104&2328&0.529 \\
    \bottomrule
    \end{tabular}
    }
    \caption{AS$^N_T$ replays. Res.=AS's result. OL=opponent level. GM=Grand-Master. M=Master. UN=Unranked. E$^A$=EPM for AS. E$^P$=EPM for player. A$^A$=APM for AS. A$^P$=APM for player. CO=camera op. AO=all op. NCR=non-camera rate.}
    \label{tab:AS Terran 1}
\end{table*}

\begin{table*}[t]
    \centering
    \scalebox{0.8}{
    \begin{tabular}{| c | l |}
    \toprule
    ID  &Description of Game Process \\
    \midrule
     01& PL (Terran) adopted Two Barracks Proxy Rush, AS lost. Although AS has scouted, the PL's 2 barracks are in covert positions, so AS didn't find it.       \\
     02& PL used Medivac to load 4 Hellion, airdropped at AS's house, and was beaten back by AS. After that, AS used an airdrop to PL's base, and PL quit.     \\
     03& PL used Cannon Rush but didn't build below the high ground, being resisted. But AS can't build sub-bases for a long time. PL gained an advantage.           \\
     04& AS's harassment didn't work. In contrast, PL's 6 Zerglings destroyed AS's sub-base, Ravagers pressed AS at home. FB: PL's 200-population won.    \\
     05& The Barrack Rush used by PL was resisted. AS uses Hellion and Banshee to sneak to gain advantages. Then PL launched an attack but was resisted.               \\
     06& PL was harassed by AS without any countermeasures. Then PL and AS competed in Macro-management. PL lost in several battles and quit.              \\
     07& PL was harassed by AS several times and suffered a great loss. PL cannot handle flexible harassment from Marine and Medivac.        \\
     08& PL uses Proxy Barrack and Reapers to suppress the sub-bases of AS. PL better defends against AS harassment and aerial harassment (by Cyclone).         \\
     09& AS's sneak was fully resisted. AS's offensive forces were eliminated by PL. AS's Medivac was pursued by air units. Finally, AS was defeated.            \\
     10& PL used Mutalisk to sneak attack and gain advantages. But AS used many Marines to forcibly attack, killing many Drones to strike PL's economy.             \\
     11& PL resisted Reaper, Hellion, and Banshee well. However, AS secretly built a sub-base in a distant place and gained an economic advantage.       \\
     12& AS harassed with 4 Hellion. PL (Protoss) didn't block the entrance and didn't immediately let the workers retreat. PL lost all workers and quit.   \\
     13& PL built a stronghold near AS's base and used 4 Tempests to destroy the AS's forces one by one. AS lost its sub-base and was starved to death.       \\
     14& In the early stage, PL gained an advantage and destroy AS's sub-base. But AS secretly built 2nd sub-base and 3rd sub-base, gaining advantages.      \\
     15& PL used Proxy Barrack Rush plus Marauder, gaining advantages. But AS produced Banshee and used SCV defend. PL lost all Marauder and quit.         \\
     16& At first, PL used Stalker Blink into AS's home and killed some Tanks. Stalker resisted Banshee in the base. PL won by some waves of attack.          \\
     17& PL used Roach Rush to suppress AS's sub-base. AS saved workers and instead killed many PL's workers with a few Marine attacks. FB: AS won.      \\
     18& PL's many workers were killed by AS's Hellion. Then more workers died by AS's Marine attack. Finally, PL's troops were defeated and quit.             \\
     19& PL was continuously harassed by AS. In a middle battle, PL handled not well and many troops were wiped out by the AS's Tanks. PL quit.      \\
     20& PL deceived AS. His sub-base is built in the AS's 3rd mine. Then PL used Ravager to suppress the AS's sub-base and finally won with Mutalisk.   \\
     21& AS's Banshee killed many PL's workers. The PL's harassment was resisted. FB: AS's 4 Tanks and many Marines wiped out the Protoss troops.          \\
     22& AS used Hellion, Raven, and Banshee to kill many PL's workers. PL has almost no effective harassment. When PL's air force was wiped, PL quit.           \\
     23& AS's Hellion killed PL's many workers. PL counter-attacked with Nydus but the effect is not good. FB: Roaches were killed many by Marines.      \\
     24& AS's Tanks, Marines, and SCVs push. PL fought and retreated, while using some troops to intercept the Terran attack, outflanked the AS troops.          \\
     25& PL's many workers were killed by Reaper, Hellion, and Banshee. PL also didn't perform well on the frontal battlefield. FB: PL lost.         \\
     26& In the early stage, AS used 4 Hellions to kill some workers of PL. But in the next few battles, PL all won.                                \\
     27& AS did not scout. AS launched 3 harassment: Reaper, Hellion, Banshee. Cloak's research was just completed when Banshee arrived at PL's base.          \\
     28& Zerg (PL) launched a wave of charges early but suffered a loss. FB: The terrain, where Zerg is located, is not good. Zerg's forces were wiped out.                 \\
     29& PL used Proxy Barracks Rush. AS found it. However, PL's Reapers suppressed the time for AS to open a sub-base, thereby gaining an advantage.        \\
     30& AS's all sneak attacks were repelled. The Medivac died one after another. AS's attack is defeated. FB: PL's Ultralisks and Zerglings defeated AS.     \\
    \bottomrule
    \end{tabular}
    }
    \caption{Description of AS$^N_T$ replays. FB=Final Battle. PL=player.}
    \label{tab:AS Terran 2}
\end{table*}

\begin{table*}[t]
    \centering
    \scalebox{0.8}{
    \begin{tabular}{c | c | c | c | l | c | c c | c c | c c c | c c c}
    \toprule
    ID &Type & OL &Res. &Strategy Order &Scout &E$^A$   &E$^P$	&A$^A$	&A$^P$  &CO$^A$   &AO$^A$	&NCR$^A$	&CO$^P$   &AO$^P$   &NCR$^P$ \\
    \midrule
    01&ZvZ& GM &Win &SubBase$-$Ravager  &none      &200	&127	&237	&309    &136&478&0.715 &480&958&0.499 \\
    02&ZvT& M & Win &SubBase$-$Ravager  &none      &212	&148	&270	&208    &419&911&0.540 &835&1736&0.519 \\
    03&ZvP& GM &Los &SubBase$-$Zergling &none      &180	&183	&216	&316    &311&823&0.622 &549&1339&0.590 \\
    04&ZvZ& M & Win &SubBase$-$Ravager  &none      &219	&146	&274	&207    &380&918&0.586 &844&1852&0.544 \\
    05&ZvZ& GM &Win &SubBase$-$Ravager  &none      &209	&204	&253	&324    &257&719&0.643 &973&1835&0.470 \\
    06&ZvP& M & Win &SubBase$-$Zergling &none      &236	&172	&298	&212    &488&1235&0.605 &1079&2469&0.563 \\
    07&ZvZ& M & Win &SubBase$-$Ravager  &none      &207	&159	&249	&328    &187&592&0.684 &417&1108&0.624 \\
    08&ZvP& M & Win &SubBase$-$Ravager  &none      &207	&159	&263	&215    &385&887&0.566 &716&1635&0.562 \\
    09&ZvZ& GM &Win &SubBase$-$Ravager  &none      &205	&130	&240	&233    &150&478&0.686 &518&929&0.442 \\
    10&ZvP& M & Win &SubBase$-$Ravager  &none      &193	&145	&240	&163    &234&626&0.626 &666&1294&0.485 \\
    11&ZvT& M & Los &SubBase$-$Ravager  &none      &195	&163	&233	&263    &385&1240&0.690 &855&2069&0.587 \\
    12&ZvZ& GM &Win &SubBase$-$Ravager  &none      &197	&172	&239	&277    &211&613&0.656 &1061&1752&0.394 \\
    13&ZvT& GM &Win &SubBase$-$Ravager  &none      &219	&135	&272	&177    &398&891&0.553 &1284&2135&0.399 \\
    14&ZvT& M & Win &SubBase$-$Ravager  &none      &211	&191	&260	&370    &303&746&0.594 &689&1728&0.601 \\
    15&ZvP& GM &Los &SubBase$-$Ravager  &none      &211	&137	&251	&165    &276&982&0.719 &782&1632&0.521 \\
    16&ZvP& GM &Win &SubBase$-$Ravager  &none      &229	&163	&272	&193    &274&744&0.632 &600&1404&0.573 \\
    17&ZvT& M & Win &SubBase$-$Ravager  &none      &232	&166	&286	&272    &705&1616&0.564 &1017&2750&0.630 \\
    18&ZvP& M & Win &SubBase$-$Ravager  &none      &184	&141	&221	&201    &250&610&0.590 &775&1398&0.446 \\
    19&ZvP& M & Win &SubBase$-$Ravager  &none      &198	&194	&239	&247    &186&551&0.662 &434&1162&0.627 \\
    20&ZvP& M & Los &SubBase$-$Ravager  &none      &132	&128	&158	&146    &130&409&0.682 &408&910&0.552 \\
    21&ZvP& GM &Win &SubBase$-$Ravager  &none      &251	&208	&317	&271    &552&1355&0.593 &1195&2831&0.578 \\
    22&ZvP& GM &Los &SubBase$-$Ravager  &none      &214	&208	&261	&333    &481&1315&0.634 &800&2313&0.654 \\
    23&ZvP& M & Win &SubBase$-$Ravager  &none      &232	&209	&288	&258    &759&2032&0.626 &1737&4548&0.618 \\
    24&ZvP& M & Los &SubBase$-$Ravager  &none      &188	&135	&228	&155    &323&1016&0.682 &813&1724&0.528 \\
    25&ZvT& M & Los &SubBase$-$Ravager  &none      &188	&149	&229	&258    &361&1015&0.644 &613&1725&0.645 \\
    26&ZvT& M & Los &SubBase$-$Ravager  &none      &197	&144	&240	&188    &1016&2084&0.512 &2341&4240&0.448 \\
    27&ZvT& UN &Los &SubBase$-$Ravager  &none      &228	&263	&277	&341    &791&1794&0.559 &1534&4144&0.630 \\
    28&ZvT& GM &Los &SubBase$-$Ravager  &none      &202	&179	&250	&211    &496&1124&0.559 &1777&3142&0.434 \\
    29&ZvZ& GM &Los &SubBase$-$Ravager  &none      &170	&201	&205	&279    &79&362&0.782 &415&955&0.565 \\
    30&ZvP& GM &Los &SubBase$-$Ravager  &none      &142	&123	&177	&148    &287&725&0.604 &860&1409&0.390 \\
    \midrule
    -&Avg& - & - & - &- &202 &166 &248 &242 &373&963&0.627 &902&1970&0.537\\
    \bottomrule
    \end{tabular}
    }
    \caption{AS$^N_Z$ replays. Res.=AS's result. OL=opponent level. GM=Grand-Master player. M=Master player. UN=Unranked player. E$^A$=EPM for AS. E$^P$=EPM for player. A$^A$=APM for AS. A$^P$=APM for player.}
    \label{tab:AS Zerg 1}
\end{table*}

\begin{table*}[t]
    \centering
    \scalebox{0.8}{
    \begin{tabular}{| c | l |}
    \toprule
    ID  &Description of Game Process \\
    \midrule
     01& PL's Zergling Rush makes no good effect. After AS built a sub-base, it used a Timing Attack (including Roach, Ravager, Queen) and defeated PL.       \\
     02& PL created no Tanks. PL's sneak attack was well resisted by AS. AS uses the Clicking way to quickly move the Camera.      \\
     03& At first, PL's worker stays in the placeholder position of AS's sub-base. AS can't build here, but order the worker to go to the 3rd mine to build it .           \\
     04& PL's Hellion's sneak attack has little effect. PL's one Battlecruiser even died during a sneak attack. FB: AS's Ravagers attack and win.     \\
     05& PL's Zergling's harassment was resisted. FB: AS's Ravager fighting against PL's Ravager, AS won due to its precise control of Roach's Biles.              \\
     06& PL's sneak attacks with Adept and Dark Templar have little effect. PL wiped out many AS's troops in a battle, but lost due to a lack of economy.              \\
     07& PL used an economical starting strategy (building a sub-base, then 3rd base). After the AS builds its sub-base, it launches a Timing Attack. PL lost.        \\
     08& After AS built its sub-base, it built the 3rd base. PL's Adept sneak attack and air drop were not effective due to poor operation. Finally, PL lost.         \\
     09& AS did a Timing attack (troops include Queen) after building the sub-base. PL knew it but still didn't replenish troops and was defeated.          \\
     10& PL did a Timing attack. However, AS divided into 3 troops, used Zergling to outflank PL on a narrow platform and Ravager to wipe out PL's troops.           \\
     11& AS built a sub-base without scouting, and the loss of being hit by PL's Proxy Barrack was great. AS did poorly in resisting Banshee sneak attack.        \\
     12& PL's Baneling's harassment has little effect. When PL owned 3 mines, AS did a Timing Attack. AS's micro-operations on Ravager defeated PL.   \\
     13& The PL's harassment has no effect. The successive waves of PL's attacks were not only ineffective, but also lost a lot of troops. AS defeated PL.       \\
     14& PL used Hellion, and Banshee to harass. However, the loss of Hellion in the mid-term is great. When AS attacked, PL have no Tanks and lost.      \\
     15& AS's sub-base was blockaded by PL with Cannon. AS's troops of Queen, Roach and Ravager fighted against PL's troops of Stalker and Immortal.        \\
     16& Protoss (PL) lost all of its Adepts in the mid-term sneak attack. Then PL was defeated by a wave of Ravager+Queen combination of AS.         \\
     17& PL's multi-line airdrops in the mid-term have a good effect. FB: PL has no Tanks switched to Siege mode, making defense broken by AS's troops.     \\
     18& PL harass without 2 Adepts. PL didn't block the entrance to its sub-base. After AS builing its sub-base, it used a Timing Attack to defeat PL.          \\
     19& After the AS builds the sub-base, it uses a mixed force composed of Zergling, Queen and Ravager to directly beat the player.  \\
     20& AS was defeated by Protoss (PL)'s Cannon Rush.  \\
     21& PL (Protoss) played steadily and faced AS in a decisive battle. The forces of AS consist of Ravager, Queen and Zergling. PL was defeated.         \\
     22& PL's handling in the early stage is not good. But PL used Prism and two Archons to attract AS troops for a long time in the mid-term.         \\
     23& PL's Cannon Rush destroyed AS's sub-base. However, AS had set up sub-bases in the 3rd and 4th mines, while PL was too slow to build it, lost.
     \\
     24& Protoss (PL) built a Cannon at AS's sub-base, causing AS's response disorder. In the mid-term, AS was attracted by 1 Prism plus 2 Immortal.         \\
     25& PL used Bunker Rush, causing AS lost 2nd and 3rd bases. Although AS's micro-operations on Ravager saved a bit of the situation, it finally lost.      \\
     26& PL's multi-line airdrop has been constraining AS. Finally, PL and AS entered into the situation of ``swapping homes''. AS lost first.                                \\
     27& PL blocked the road with Supply Depot. AS doesn't know how to get in . In the mid-term of the game, AS has the advantage, but has not attacked.          \\
     28& AS was about to win at the beginning, but was harrassed by a PL's Battlecruiser. AS's workers and Overlord suffered a great loss. AS finally lost.                \\
     29& AS didn't scout, directly building a sub-base. PL used 6 Zergling+6 workers to Rush, and workers built Spine Crawler in AS's base. Game ended.        \\
     30& PL used Cannon Rush at the AS's sub-base. AS didn't know how to handle it. Its workers went around and many of them were killed. AS lost.    \\
    \bottomrule
    \end{tabular}
    }
    \caption{Description of AS$^N_Z$ replays. FB=Final Battle. PL=player.}
    \label{tab:AS Zerg 2}
\end{table*}

\clearpage
\newpage

\section{Analysis of All Replays}
The analysis of replays gives our discovery from the replays in Table~\ref{tab:AS Protoss 3},~\ref{tab:AS Terran 3}, and~\ref{tab:AS Zerg 3}. If the replay's analysis content is ``None'', it means there is no special thing worth saying in that replay.

\begin{table*}[t]
    \centering
    \scalebox{0.8}{
    \begin{tabular}{| c | l |}
    \toprule
    ID & Analysis of the Game \\
    \midrule
    01  & None.                                \\
    02  & We find AS wins, relying on frontal battles and micro-operations; AS loses due to a sneak attack, suppressed by tactics, etc.            \\ 
    03  & AS is totally unaware to prevent a sneak attack from Banelings. The player killed 9 workers of AS with two Banelings.          \\
    04  & When the Disruptor killed 4 Ravagers in one shot, PL's mentality seemed to give up. There are 3 Disruptors in the AS troops.          \\
    05  & When HP attacked at the final battle, AS immediately established two Shield Battery in the sub-base, and let workers also join in defend. \\
    06  & None.          \\
    07  & None.                \\
    08  &  AS saw three Zerg workers coming out of the base, but AS did not realize that it would be attacked by the opponent through Crawler Rush.    \\
    09  & None.     \\
    10  & We find AS's harassment doesn't care about the loss. If it can kill workers, it will be pleased to use its battle unit to exchange.       \\
    11  & None.                     \\
    12  & None.                                 \\
    13  & AS's offensive forces include Disrupter and Warp Prism, and use them effectively. AS's third base built a lot of Cannon to defend.                  \\
    14  & AS is also experienced in preventing Nydus Worm's fast Rush. AS can even use workers to resist Zerglings attacking.        \\
    15  & The instantaneous EPM of AS reached 360, which was three times that of GM players.             \\
    16  & AS's sneak attack was excessively ferocious.               \\
    17  & In FB, AS has superbly used Warp Prism, and many attacks have been evaded by Prism's loading, making AS feels like a scripted bot.         \\
    18  & The 2 Oracles operated by the AS can instantly kill one worker of PL. This operation seems to be much more powerful than PL.               \\
    19  & AS uses load and unload to control the collaboration between Prism and Immortal, so as to maximize the use of Prism.           \\
    20  &  AS used Adept and Oracle to destroy PL's workers without caring about loss. As long as the costs of the two are close, AS has advantages.         \\
    21  &  AS builds many sub-bases, and builds Cannons in each sub-base to defend. The PL's Zerg (PL) economy is not even as good as the Protoss.        \\
    22  &  AS's manipulation of Prism gives it an advantage in battle. When AS uses Phoenix, it performs very well. \\
    23  &  AS's mixed troops include Disruptor, Dark Templar, Warp Prism, and Archon.            \\
    24  &  This PL's Cannon Rush has performed very well.                   \\
    25  &  Although PL is unranked, it is definitely at the level of GM. AS is slightly inferior in all aspects of operation in the face of PL.          \\
    26  & PL wanted to just use Stalker to defend against the harassment of AS's Oracle, but the effect was not good.                   \\
    27  & The micro-operation of AS is too strong, so the advantages gained are too great.           \\
    28  & AS lacks a reasoning module. Its victory mainly relies on the imitation of humans, exaggerated precision, and an unchanging style of play.                \\
    29  & The operation used by AS to defend Cannon Rush obviously has more room for optimization.           \\
    30  & When using Adept's skills, AS needs not move the camera to the skill's target position, meaning AS uses raw action and no screen information. \\
    \bottomrule
    \end{tabular}
    }
    \caption{Analysis of AS$^N_P$ replays. FB=Final Battle. PL=player.}
    \label{tab:AS Protoss 3}
\end{table*}

\begin{table*}[t]
    \centering
    \scalebox{0.8}{
    \begin{tabular}{| c | l |}
    \toprule
    ID & Analysis of the Game \\
    \midrule
    01  & AS does not scout, except in TvT it scouts the wild near home. If AS goes to PL's home, it may be able to discover PL's tactical intent.                               \\
    02  & During TvT, AS will scout around the base to see if the opponent will adopt Proxy Barrack Rush.            \\ 
    03  & PL (Protoss) released a very beautiful Psionic Storm several times in the final battle and gained a great advantage.          \\
    04  & When the Disruptor killed 4 Ravagers in one shot, PL's mentality seemed to give up. There are 3 Disruptors in the AS troops.          \\
    05  & PL performs not good at defending against invisible units (Banshee). He did not build any Missile Turret. \\
    06  & None.          \\
    07  & PL has almost no effective harassment to AS. After a battle in the mid-term, PL surrendered when still having a lot of minerals.               \\
    08  & PL's last few waves of offense were also very decisive. This unranked player seems to be a Grand-Master player.   \\
    09  & Very good GM player. This is the level of a real Zerg GM player!     \\
    10  & PL played well, but it lost a lot when resisting Hellion. If PL's Mutalisk can continue to harass the AS's base, it should be able to win.       \\
    11  & The troops on both sides died in a battle. But AS's economy was not affected. PL lost a sub-base and economic income, and eventually lost.                     \\
    12  & This Master level players seems so amateur.                                \\
    13  & When confronted with these novel tactics, AS didn't have a counterattack strategy, and was beaten very embarrassed (due to no scouting).                  \\
    14  & PL wants to compete with AS on Macro-management. However, AS harassed, while using the main force to attack. PL lost after a battle.        \\
    15  & It is worth noting that AS initially sent workers to scout near its third mine. AS saw the PL's barracks and immediately built Bunker.             \\
    16  & As PL blocked the entrance with buildings, Hellion of AS kept wanting to go in and turned into spinning outside. This scene looks funny.               \\
    17  & On the battlefield, when the PL's Baneling rushed forward, they were evaded by the AS troops many times.         \\
    18  & When Hellion attack, PL's operation felt as if it had given its workers to AS. Looking at the whole game, the PL has no level of GM at all.               \\
    19  & At the time of surrender, PL's strength can still continue to fight, but his confidence collapsed too fast and chose to surrender.            \\
    20  &  PL used Queen + Spine to resist harassment. AS has always wanted to build a base on Creep. A very exciting battle shows the wisdom of mankind.         \\
    21  &  AS's combination of Tank and Marine performed well in pushing, and it played a role in attacking Protoss.       \\
    22  &  PL gave up the game when his population was similar to that of AS. \\
    23  &  PL did not have an effective way to fight against a cluster of Marines, and AS used Marines to kill many PL's units.            \\
    24  &  PL played patiently. His one force eliminated reinforcements from AS's home, and used 2 troops to outflank AS's main force to win the battle.                   \\
    25  &  PL used Hydralisks to attack Marines and Tanks, but the effect was not ideal. Mutalisk should be a better choice in this situation.          \\
    26  & AS's control of Medivac is a bit weak, and several troops were died in vain. AS has wanted to build buildings on Zerg Creep several times.                   \\
    27  & PL doesn't t block its entrance or build Cannons. PL's 9 workers were killed by AS's Hellion. After seeing Banshee invisible, PL gave up.           \\
    28  & The control of the few waves of PL's battles were too casual. Its troops shouldn't have suffered such a big loss in these battles.                \\
    29  & AS secretly built a sub-base far but was discovered by PL. When AS sent workers there, they were killed (no retreating) by PL's troops on the way.           \\
    30  & PL's operation and awareness are very strong. It was later learned that this Zerg player is Sierra, the world's top player so far. \\
    \bottomrule
    \end{tabular}
    }
    \caption{Analysis of AS$^N_T$ replays. FB=Final Battle. PL=player.}
    \label{tab:AS Terran 3}
\end{table*}

\begin{table*}[t]
    \centering
    \scalebox{0.8}{
    \begin{tabular}{| c | l |}
    \toprule
    ID & Analysis of the Game \\
    \midrule
    01  & The strength of the GM PL is not strong. Since levels of GM players are quite different, the fact AS has reached the GM level doesn't explain much.                               \\
    02  & AS needs no selection when operating units. AS uses a lot of Ravagers to gain advantages by quickly and continuously releasing Corrosive Bile.            \\ 
    03  & When AS wants to build Roach Warren, the placeholder is occupied by another worker, and construction fails. In all the game, AS didn't rebuild it.          \\
    04  & When fighting against Tanks, AS will use Ravager's Bile plus Queen's Transfusion for micro-manipulation and suppression.          \\
    05  & When using Ravager, PL made a fatal mistake: several Corrosive Biles were released on an empty place, making the battle lost. \\
    06  & None.          \\
    07  & When AS attacks, PL can defend with Zerglings at first. But when AS's Queens also come to the battlefield, PL can't resist AS's mixed forces.               \\
    08  & Since PL didn't block its entrance, AS's Zerglings rushed in and killed many PL's workers. There is a gap between the levels of the two.   \\
    09  & The GM PL has no sense of crisis. He placed Roach Warren on the outside which was the first to be destroyed, making it hard to replenish troops.     \\
    10  & The timing of retreat and attack of AS's several waves of troops is appropriate. We think this is an impressive and excellent battle fought by AS.       \\
    11  & This replay highlights the drawbacks of AS's Zerg agent not scouting and not defending against air attacks by invisible units (Cloaked Banshee).                     \\
    12  & AS will transform a Roach into a Ravager when it reaches low health.                               \\
    13  & This GM's level is very watery. At first, troops were sacrificed in vain. Then Thors attack AS without manipulation and died (incomprehensible).                 \\
    14  & This human Terran player does not use tanks when defending. We feel that he underestimated the opponent too much.       \\
    15  & PL choose unique tactics, and then use micro-operations to compete with AS. The PL did not fall behind and won in the end.            \\
    16  & None.            \\
    17  & PL played well. He often uses multi-line airdrops. This PL should have the level of GM.      \\
    18  & None.             \\
    19  & None.       \\
    20  & It seems that the Zerg agent of AS has less training against Cannon Rush.         \\
    21  & The Zerg of AS likes Ravager just like the Protoss of AS$^B$ likes Stalker. This GM feels rather amateurish, and the details are not handled well.       \\
    22  & Strategy of PL is similar to MaNa's strategy on the Live show. He used Dark Templar and Prism to attract the AS's Zerg, and won in the last battle.  \\
    23  & PL had an advantage, but lost due to carelessness. In middle-term, AS divided its troops and outflanked PL's troops, causing PL to suffer a loss.         \\
    24  & AS's economy hadn't been good. Its sub-base was suppressed by Cannon. After that, AS was attracted by PL's Prism plus Immortal for a long time.                \\
    25  & Later, when AS built a sub-base in the third mine, it was defeated by a wave of Timing Attack from PL.    \\
    26  & AS isn't good at playing in the situation of ``swapping homes''. Although AS had more troops, it lost due to its buildings are being destroyed first.                 \\
    27  & We find the reason for AS's not attacking is: enemy's base is unknown (road blocked and AS didn't use Overlord scout). So its troops go around.           \\
    28  & When facing novel tactics (such as Battlecruiser), AS doesn't know how to deal with it. In the end it was defeated by a bunch of Battlecruiser.                \\
    29  & This replay embodies the drawbacks of AS's non-reconnaissance. AS was directly defeated when facing the opponent's Rush.           \\
    30  & When AS is suppressed by PL in the sub-base of the 2nd mine, AS didn't know how to play, indicating AS's ability to respond to changes is poor.\\
    \bottomrule
    \end{tabular}
    }
    \caption{Analysis of AS$^N_Z$ replays. FB=Final Battle. PL=player.}
    \label{tab:AS Zerg 3}
\end{table*}

\end{appendices}

\end{document}